%% file: emnlp2020.tex
\documentclass[11pt,a4paper]{article}
\usepackage[hyperref]{emnlp2020}
\usepackage{enumitem}
\usepackage{times}
\usepackage{latexsym}
\usepackage{graphicx}
\usepackage{url}
\usepackage{amsmath}
\usepackage{float}
\usepackage{xcolor}
\usepackage{subcaption}
\usepackage{multirow}
\usepackage{amsfonts}
\usepackage{booktabs}
\usepackage{bbm}
\usepackage{pifont}
\usepackage{makecell,rotating}
    \setcellgapes{5pt}

\hypersetup{colorlinks,urlcolor={blue}}

\newcommand*{\affmark}[1][*]{\textsuperscript{#1}}
\newcommand\blfootnote[1]{%
  \begingroup
  \renewcommand\thefootnote{}\footnote{#1}%
  \addtocounter{footnote}{-1}%
  \endgroup
}

\usepackage{microtype}

\aclfinalcopy 
\def\aclpaperid{7} 

\newcommand{\footnoteref}[1]{\textsuperscript{\ref{#1}}}

\title{GenAug: Data Augmentation for Finetuning Text Generators}

\author{Steven Y. Feng\bf{\thanks{\quad Equal contribution by the two authors.}}, \affmark[1] Varun Gangal\footnotemark[1], \affmark[1] Dongyeop Kang,\affmark[2] Teruko Mitamura,\affmark[1] Eduard Hovy\affmark[1]\\

  \affmark[1]Carnegie Mellon University\\
  {\tt \{syfeng,vgangal,teruko,hovy\}@cs.cmu.edu }\\
  \affmark[2]University of California, Berkeley\\
  \texttt{dongyeopk@berkeley.edu} 
}
\date{}

\begin{document}
\maketitle
\setlength{\abovedisplayskip}{4pt}
\setlength{\belowdisplayskip}{4pt}

\begin{abstract}
\input{sections/abstract.tex}
\end{abstract}

\section{Introduction}
\label{sec:intro}
\input{sections/intro.tex}

\section{Methodology}
\label{sec:methodology}
\input{sections/methods.tex}

\section{Experiments}
\label{sec:experiments}
\input{sections/experiments.tex}

\section{Results and Analysis}
\label{sec:results_and_analysis}
\input{sections/results_and_analysis.tex}

\section{Related Work}
\label{sec:related}
\input{sections/related-work.tex}

\section{Conclusion and Future Work}
\label{sec:conclusions}
\input{sections/conclusions.tex}
\section*{Acknowledgments}
We thank the three anonymous reviewers for their comments and feedback.

\bibliographystyle{acl_natbib}
\bibliography{anthology,emnlp2020}

\newpage
\appendix
\section*{Appendices}
\input{sections/appendix.tex}

\end{document}

%% file: sections/abstract.tex


In this paper, we investigate data augmentation for text generation, which we call \textit{GenAug}. Text generation and language modeling are important tasks within natural language processing, and are especially challenging for low-data regimes. We propose and evaluate various augmentation methods, including some that incorporate external knowledge, for finetuning GPT-2 on a subset of Yelp Reviews. We also examine the relationship between the amount of augmentation and the quality of the generated text. We utilize several metrics that evaluate important aspects of the generated text including its diversity and fluency. Our experiments demonstrate that insertion of character-level synthetic noise and keyword replacement with hypernyms are effective augmentation methods, and that the quality of generations improves to a peak at approximately three times the amount of original data.

%% file: sections/intro.tex
Text generation is an important but difficult task within natural language processing (NLP). A major goal is for dialogue agents to generate human-like text. The development of strong pretrained text generators like GPT-2 \cite{radford2019language} has made it easier to perform generation for new domains or task specifications. These models are typically finetuned on downstream tasks such as classification; however, the first stage of their training is language modeling. Effective language models  are important not only for generation but many NLP tasks.

\begin{table}
\begin{tabular}{@{}ll@{}}
\includegraphics[width=0.47\textwidth]{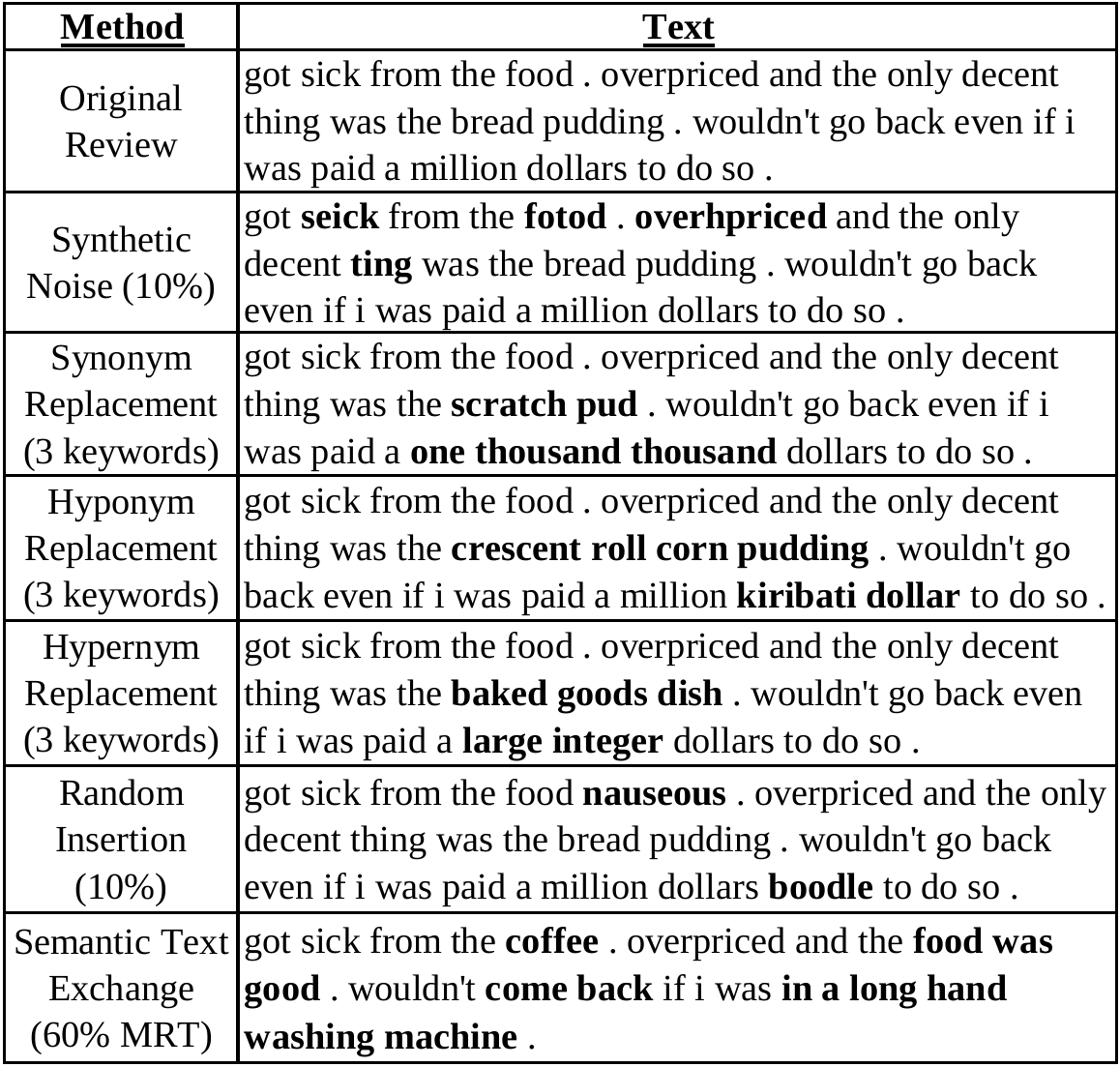}\\
\end{tabular}
  \caption{\label{tab:qualitative_1} Example of a Yelp review and its variations using our augmentation methods. Changes are bolded.}
\end{table}

In-domain examples are needed for finetuning. Otherwise, the generated text, though fluent English, will not faithfully imbibe domain properties such as the vocabulary preferred, domain shifts in word meaning, and domain distribution over properties such as sentiment. The learned language model will also poorly replicate the domain. However, many domains are low-data. These models do not have enough data to learn domain-specific aspects of the text, especially without sacrificing aspects such as its fluency and diversity.

One approach is with text data augmentation. There is constantly an increasing demand for large amounts of text data. Compared to fields such as computer vision, augmentation techniques for NLP are limited. Collecting and cleaning data manually requires time and effort. Also, certain domains do not have sufficient data available to begin with.\blfootnote{Code: \url{https://github.com/styfeng/GenAug}}

Prior work in text augmentation has focused on classification tasks, and there has been limited investigation for generation. A possible explanation is that generation is more complicated; rather than predicting the correct label, the text itself must be produced and should satisfy properties typical of human text such as being fluent, logical, and diverse. Evaluation of the text is also more difficult.

In this work, we focus on data augmentation for text generation. We call this \textit{GenAug}, and to the best of our knowledge, are the first to investigate it. We explore various augmentation methods such as semantic text exchange (STE)~\cite{feng2019keep} and replacing keywords within examples from a small subset of the Yelp Reviews dataset \cite{YelpReviewsDataset}. See Table~\ref{tab:qualitative_1} for examples.\footnote{See Appendix \S\ref{sec:appendix_augmentation_variation_examples} for more augmentation examples.} We also assess the impact of augmentation amount: from 1.5x to 4x the original amount of training data. 

We evaluate the quality of generated text by GPT-2 after finetuning on our augmented data compared to the original data only. We illustrate that several augmentation methods improve the quality of the generations. We also show that the quality follows a trend with the augmentation amount: it increases until a peak and decreases thereafter. Overall, our major contributions can be summarized as follows:

\begin{itemize}[noitemsep,topsep=2pt]
  \item We propose \textit{GenAug}, which is data augmentation specifically for text generation.
  \item We introduce and evaluate various augmentation methods for GenAug including inserting synthetic noise and integrating external knowledge through lexical databases for keyword replacement. We demonstrate that synthetic noise and replacement with hypernyms improve the quality of generations.\footnote{\label{note1}See Section \S\ref{sec:results_and_analysis} for results and analysis.}
  \item We investigate the effects of the augmentation amount and discover that performance improves until approximately three times the original training data, where all aspects of the generated text are noticeably improved upon.\footnoteref{note1}
  \item We propose and use a mix of new and existing metrics for evaluating aspects of the text including its diversity, fluency, semantic content preservation, and sentiment consistency.\footnote{See Section \S\ref{sec:metrics} for evaluation metrics.}
\end{itemize}

%% file: sections/methods.tex
\subsection{Model: GPT-2}
\label{sec:models}
We use OpenAI's GPT-2 \cite{radford2019language}, specifically its default pretrained model with 117M parameters. GPT-2 is a large transformer-based language model trained to predict the next word given previous words in a text. It is trained on \textit{WebText} - a variety of internet data from sources such as Reddit, and has been shown to generate fluent text given different input prompts.

We choose this model as it is reasonably sized, frequently used as a pretrained text generator, and would thus benefit significantly from our experiments and analysis. We use HuggingFace's implementation of GPT-2 \cite{Wolf2019HuggingFacesTS}.

\subsection{Dataset: Yelp Reviews (YR)}
\label{sec:datasets}
The Yelp Reviews (YR) dataset contains user reviews on businesses. We choose YR as it differs substantially in domain from the ``\textit{WebText}" data used to train GPT-2, which consisted mainly of newswire and discussion forum threads. Unlike other review corpora such as SST-2 \cite{socher2013recursive}, YR contains long reviews with many sentences, making generation non-trivial. 

We randomly select a small subset of YR for our experiments: a training split of 50K, validation split of 15K, and test split of 2K. This is approximately 1\% of YR, replicating a low-data regime. We call this \textit{Yelp-LR} or \textit{YLR} (LR stands for low-resource). We include a proportion of reviews of each star rating equal to the proportions within YR to replicate the distribution of sentiment in YR.\footnote{See Section \S\ref{sec:processing} for preprocessing details for this dataset.}

Finetuning GPT-2 on YLR represents the gold or baseline model. For each augmentation experiment, we combine YLR with our augmented data and finetune GPT-2 on this combination while using the same 15K validation and 2K test splits.

\subsection{Text Augmentation Methods (AM)}
\label{sec:variations}
We explore various augmentation methods (AM) to produce different versions of our training reviews\footnote{We also tried syntactic paraphrasing using SCPNs \cite{wieting-gimpel-2017-revisiting} but found the paraphrase quality poor and hard to control for meaning preservation and fluency.} (see Table~\ref{tab:qualitative_1} for examples), and analyze their effects on GPT-2's generations. We split each review in half; a prompt and a continuation portion. We finetune GPT-2 on the entire reviews, but different AM are applied to either the prompt portion or entire review. We feed the prompt portion of test reviews as input to generate continuations.

\subsubsection{Random Insertion, Deletion, \& Swap}
We experiment with random insertion, deletion, and swap (the \textit{``Random Trio"}) on our entire reviews. \citet{wei2019eda} used these along with synonym replacement for text classification, and we investigate their performance for generation. 

For each training example, we randomly swap the positions of two words, insert a random synonym of a word that is not a stopword\footnote{We use the stopwords list from~\citet{stopwords}.} into a random location, and remove a word, with $\alpha$ = 5\% and 10\% (5\% and 10\% of the words are changed). Hence, we produce six total variations per example.

\subsubsection{Semantic Text Exchange (STE)}
\label{subsection:STE}
We investigate Semantic Text Exchange (STE) as introduced in \citet{feng2019keep} on the entire reviews. STE adjusts text to fit the semantic context of a word/phrase called the replacement entity (RE). We use \citet{feng2019keep}'s SMERTI-Transformer by training on a subset of YLR.\footnote{See Section \S\ref{sec:SMERTITT} for SMERTI training details.} It inserts the RE into the text by replacing another entity, masks words similar to the replaced entity, and fills in these masks using a masked language model.

SMERTI is designed for shorter text due to the limited ability of the model to learn longer temporal dependencies.\footnote{\citet{feng2019keep} perform STE on text 	$\leq$ 20 words long.} We break each review into windows, and perform STE on each. Our augmentations are the concatenation of the semantically adjusted windows. For each window, a random RE is chosen. The candidates REs are 150 of the 200 most frequent nouns in SMERTI's training set.\footnote{See Appendix \S\ref{sec:appendixWindowDetails} for sliding window algorithm details.} We use masking rate thresholds (MRT) of 20\%, 40\%, and 60\% 
, which represent the maximum proportion of the text that can be masked and replaced.

\subsubsection{Synthetic Noise}
We add character-level synthetic noise to the prompt portion of reviews. For every word, at every character, we perform a character insertion, deletion, or swapping of two side-by-side characters. The insertions are lowercase letters. 

The three events have an equal chance of occurring at every character equal to one-third the overall level of noise. We ignore the first and last character of every word to more closely imitate natural noise and typos \cite{belinkov2017synthetic}. We produce 5\%, 10\%, and 15\% noise variations per review. The noised prompt is combined with the original continuation to form the augmentations.

\subsubsection{Keyword Replacement}
We experiment with replacing keywords within entire reviews. We use RAKE \cite{rose2010automatic} for keyword extraction. Candidate replacements are extracted from the lexical database WordNet \cite{miller1998wordnet}. We replace up to three keywords for each review, resulting in a maximum of three augmentations for each review. Unlike STE, our goal is not to adjust the text's overall semantics.

The keywords replaced are ordered by their RAKE score (e.g. the probability of being a keyword) and replaced with words with the same overall part-of-speech (POS). We use the Stanford POS Tagger~\cite{stanfordPOStagger}. Previous replacements are kept intact as further ones occur. There are three replacement methods:
\begin{enumerate}[noitemsep,topsep=2pt]
\item {\bf Synonyms Replacement (WN-Syns)} replaces each chosen keyword with a randomly chosen synonym of the same POS, preserving the text's semantics as much as possible.
\item {\bf Hyponyms Replacement (WN-Hypos)} replaces each chosen keyword with a randomly chosen hyponym of the same POS that has more specific meaning. Words can have multiple hyponyms which differ semantically.
\item {\bf Hypernyms Replacement (WN-Hypers)} replaces each chosen keyword with the closest (lowest) hypernym of the same POS that carries more broad and high-level meaning.
\end{enumerate}
\vspace{-0.35\abovedisplayskip}
\textbf{\subsection{Text Augmentation Amounts}}
\label{sec:amounts}
We also assess the impact of the amount of augmentation on the generated text. Specifically, \textbf{1.5x}, \textbf{2x}, \textbf{3x}, and \textbf{4x} the original amount of data (e.g. 4x refers to each example having three augmentations). We use a combination of synthetic noise, STE, and keyword replacement, each augmenting $\frac{1}{3}$ the YLR training examples (WN-Syns, Hypos, and Hypers each augment $\frac{1}{9}$).
\\

\subsection{Evaluation Metrics}
\label{sec:metrics}
We evaluate generated continuations using various metrics assessing major aspects of the text including its diversity, fluency, semantic content preservation, and sentiment consistency. Arguably the two most important are text fluency and diversity.\footnote{We do not use BLEU~\cite{papineni2002bleu} as we only have a single ground-truth continuation per review.}

\subsubsection{Diversity}
We pick a broad range of diversity measures for both intra- and inter-continuation diversity.\footnote{We evaluate diversity on the generated continuations only (not concatenated with their corresponding prompts).}
\begin{enumerate}[noitemsep,topsep=2pt]
    \item \textsc{Self-BLEU (SBLEU) \cite{selfbleu}}, for a sample population $S$, measures the mean similarity of each sample to other samples. It is expressed as $E_{s \sim S }[BLEU(s,S-\{s\})]$, where $BLEU(h,R)$ is the BLEU-4 score of a hypothesis $h$ measured against a set of references $R$. We measure the average SBLEU of every batch of 100 continuations per test prompt.\footnote{This is because we generate 100 continuations per test example. See Section \S\ref{sec:setup} for more.} Lower SBLEU values represent higher inter-continuation diversity.
    \item \textsc{Unique Trigrams (UTR)} \cite{tevet2020evaluating,li2015diversity} measures the ratio of unique to total trigrams in a population of generations. Higher UTR represents greater diversity. Since UTR is defined at the population level, it can assess the extent of cross-continuation repetition.
    \item \textsc{Type-Token Ratio (TTR)} is the ratio of unique to total tokens in a piece of text, and serves as a measure of intra-continuation diversity. The higher the TTR, the more varied the vocabulary in a continuation.
    \item \textsc{Rare-Words} (\textsc{RWords}) \cite{see2019massively} is defined by the following:
    \begin{align*}
    E_{s \sim S }[\sum_{w \in s} -\log \frac{n_{train}(w)}{N_{train}}] 
    \end{align*}
    where $n_{train}(w)$ and $N_{train}$ are the corpus frequency of word $w$ and the total corpus word count, respectively. Our corpus here is the 50K YLR training split. Lower values indicate usage of more rare words (less frequent in the corpus) and higher diversity.
\end{enumerate}

\subsubsection{Fluency}
Fluency, also known as naturalness or readability, is a measure of how fluent text is. The higher the fluency, the more it imitates grammatically and logically correct human text.\footnote{We evaluate perplexity and SLOR on the concatenations of the generated continuations with their corresponding prompts, and Spellcheck on the generated continuations only.} 
\begin{enumerate}[noitemsep,topsep=2pt]
    \item \textsc{Perplexity (PPL)} is defined as:
    \begin{align*}
    PPL(S)&=exp(-\frac{1}{|S|}ln(p_M(S)))
    \end{align*}
    where $S$ is a piece of text and $p_M(S)$ is the probability assigned to $S$ by the language model. We finetune GPT-2 on a two-million review subset of YR (with a 500K additional validation split) and use this finetuned model for PPL evaluation. Outputs less likely to be seen in YR will typically have higher PPL.
    \item \textsc{SLOR} (syntactic log-odds ratio) \cite{kann-etal-2018-sentence} is our main fluency metric. It modifies PPL by normalizing for individual tokens (e.g. ``Zimbabwe" is less frequent than ``France" but just as fluent), and serves as a better measure. Higher SLOR represents higher fluency. The equation for SLOR is as follows: 
    \begin{align*}
      SLOR(S)&=\frac{1}{|S|}(ln(p_M(S))-ln(\prod_{t\in S}p(t)))
    \end{align*}
    where $|S|$ is the length of $S$ (in tokens), $p_M(S)$ is the probability of $S$ under language model $M$, and $p(t)$ are the unconditional probabilities of individual tokens (or unigrams) $t$ in $S$. We use the same finetuned GPT-2 model on YR as for PPL mentioned above for SLOR. We use the proportional frequencies of unigrams in the two-million reviews as the unconditional unigram probabilities. Specifically, for tokens $t$: $p(t)=\frac{f(t)}{z+1}$, where $f(t)$ is the frequency of token $t$ and $z = {\sum_{t} f(t)}$. 
    \item \textsc{Spellcheck:} For synthetic noise, we measure two spelling related metrics:
    \begin{enumerate}[noitemsep,topsep=2pt]
    \item \textsc{SpellWords:} average number of misspelled words per continuation.
    \item \textsc{SpellChars:} average number of character level mistakes per continuation.
    \end{enumerate}
    
    These approximately measure how \textit{noisy} the generations are, which can misleadingly improve diversity metrics. We use SymSpell~\cite{symspell2019}, which uses a Symmetric Delete Algorithm to quickly compute edit distances to a predefined dictionary. We set \textit{verbosity} to \textit{top}, 
    a prefix length of ten, and consider a maximum edit distance of five.
\end{enumerate}

\subsubsection{Semantic Content Preservation (SCP)}
SCP assesses how closely each generated continuation (hypothesis) matches in semantic content to the ground truth distribution of continuations (reference). Since the latter is unavailable in this case, we use the prompt itself as a proxy for reference.\footnote{Datasets with multiple continuations per prompt are rare, and one continuation would be insufficient in most cases.}

We use what we call the Prompt-Continuation BertScore (\textsc{BPRO}). BPRO computes average BertScore \cite{zhang2019bertscore} between each continuation and the prompt. BertScore computes per-token BERT representations for both hypothesis and reference and aligns each hypothesis token to a reference token. We prefer BertScore over symbolic measures (e.g BLEU) since it does not rely on exact string matching alone and allows soft matches between different parts of the input pair.


\subsubsection{Sentiment Consistency}
We finetune a BERT \cite{devlin2018bert} sentiment regressor on YLR by converting review stars into values between 0 and 1, inclusive, with higher values representing more positive sentiment.\footnote{See Appendix \S\ref{sec:appendix_sentiment_regressor_finetune} for regressor finetuning details.} We run the regressor on the ground-truth test reviews and the concatenation of our generated continuations with their corresponding prompts. We measure:
\begin{enumerate}[noitemsep,topsep=2pt]
    \item \textsc{SentStd:} average standard deviation of sentiment scores among each batch of 100 continuations (each concatenated with the input prompt) for a given test example. We do this for all 2000 test examples (100 prompt + continuation concatenations each) and take the average of the standard deviation values for each. A lower value indicates more consistent (lower spread) of sentiment, on average, among the continuations for each prompt.
    \item \textsc{SentDiff:} average difference in sentiment score between each batch of 100 continuations (each concatenated with the single input prompt) and the corresponding ground-truth review in its entirety (essentially, the input prompt concatenated with the ground-truth continuation). We run this for all 2000 test examples (100 prompt + continuation concatenations each) and take the average of the differences. A lower value indicates sentiment of the continuations that, on average, more closely aligns with the ground-truth reviews.
\end{enumerate}

%% file: sections/experiments.tex
\subsection{GPT-2 Finetuning}
\label{sec:finetuning}
We finetune GPT-2 with a batch size of two. We try three different learning rates on YLR: 5e-4, 5e-5, and 5e-6, and find 5e-5 results in the lowest validation perplexity and use it for all experiments. We ensure the same hyperparameters and settings are used for each experiment. Final models correspond to epochs with the lowest validation perplexity.\footnote{See Appendix \S\ref{subsec:finetuned_model_details} for details of the finetuned models.}

\subsection{SMERTI-Transformer Training}
\label{sec:SMERTITT}
We take a 25K subset of YLR's training split and a 7.5K subset of YLR's validation split. These serve as SMERTI's training and validation splits, respectively. This replicates the low-data regime, ensures SMERTI does not see additional data, and ensures SMERTI only learns from a portion of the data to prevent overfitting and repetition. 

Each chosen example is split into chunks (or windows) of up to 30 tokens each,\footnote{See Appendix \S\ref{sec:appendixWindowDetails} for sliding window algorithm details.} resulting in 144.6K total training and 43.2K total validation examples for SMERTI. We mask 20\%, 40\%, and 60\% of the words in $\frac{1}{3}$ of the examples each. We train SMERTI on this data and find the best performance after 9 epochs with a validation loss of 1.63. We use scaled dot-product attention and the same hyperparameters as \citet{feng2019keep}.\footnote{See Appendix \S\ref{sec:appendixSMERTITTDetails} for further SMERTI training details.}
\subsection{Data Processing}
\label{sec:processing}
For Yelp preprocessing, we filter out reviews that are blank, non-English, or contain URLs. For remaining ones, we remove repeated punctuations and uncommon symbols. For postprocessing, we noticed that many GPT-2 generations included trailing exclamation marks. We stripped these if more than four occurred in a row. Resulting blank continuations (very small portion of the total) were represented with a $<$\textit{blank}$>$ token and ignored during evaluation of most metrics. 

\subsection{Experimental Setup}
\label{sec:setup}
For the separate method experiments, we choose one augmentation for each training example, for a total of 2x the amount of original data. Since each method has multiple variations per training example, we randomly select one of these for each.

For the augmentation amount experiments, we ensure that larger amounts are supersets of smaller amounts - e.g. 3x contains all of the augmentation examples within 2x, and so forth.

We generate 100 continuations per test example by feeding in the prompt portions (first 50\% of words). We use the default end-of-text token, a nucleus sampling budget \cite{holtzman2019curious} of 0.9, and a length limit of 500 for the generations.

For all experiments, we run two sets of random seeds, where each set $\{rs_1, rs_2\}$ consists of $rs_1$: a seed for data preparation and selection, 
and $rs_2$: a seed for GPT-2 finetuning and generation. 
Our final evaluation results are the average results.

%% file: sections/results_and_analysis.tex
\subsection{Evaluation Results}
\begin{figure}[ht]
\begin{subfigure}{.97\columnwidth}
    \centering
    \includegraphics[width=0.99\columnwidth]{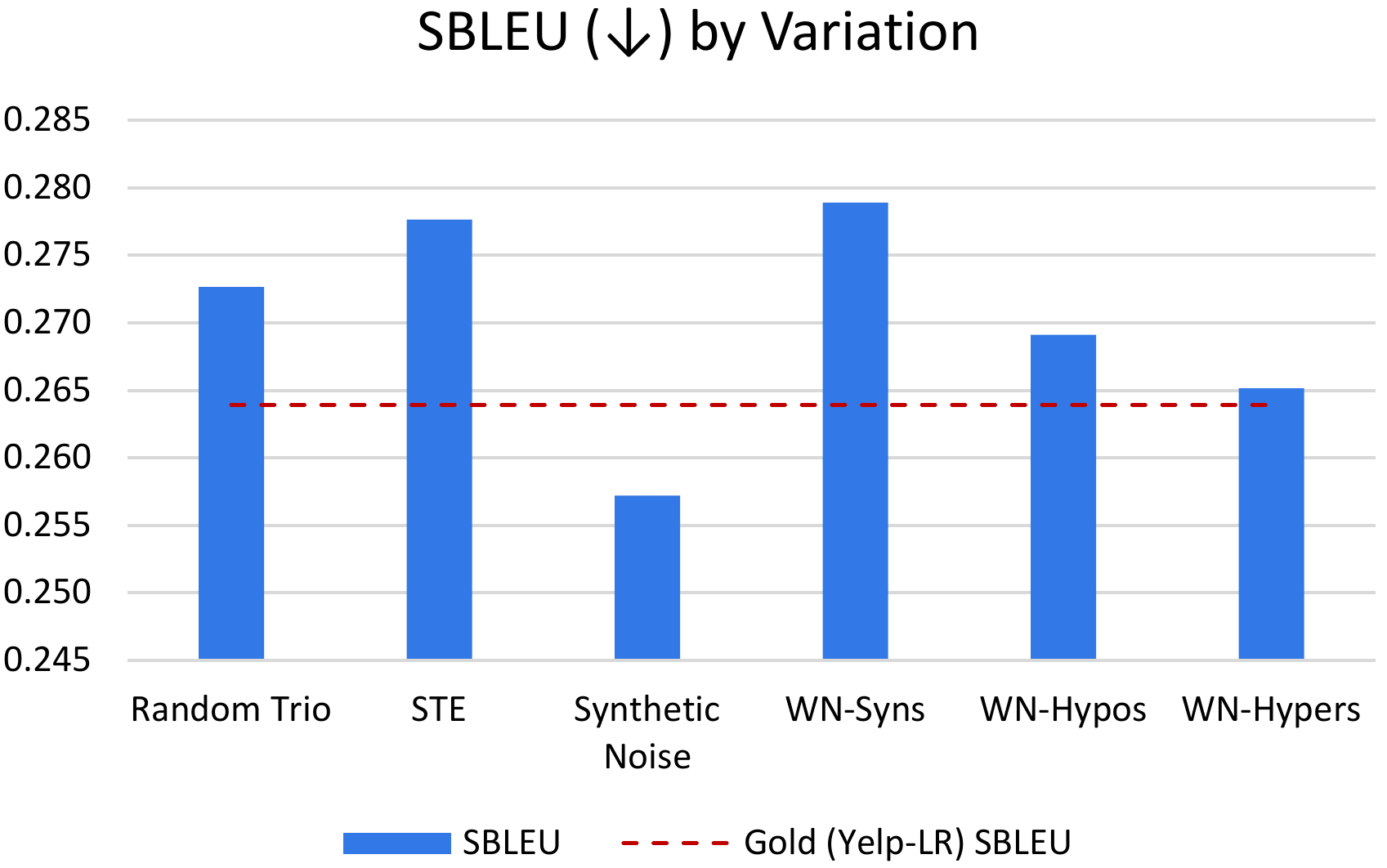}
    \caption{}
    \label{fig:variation_SBLEU}
\end{subfigure}\\
\bigskip
\begin{subfigure}{.97\columnwidth}
    \centering
    \includegraphics[width=0.99\columnwidth]{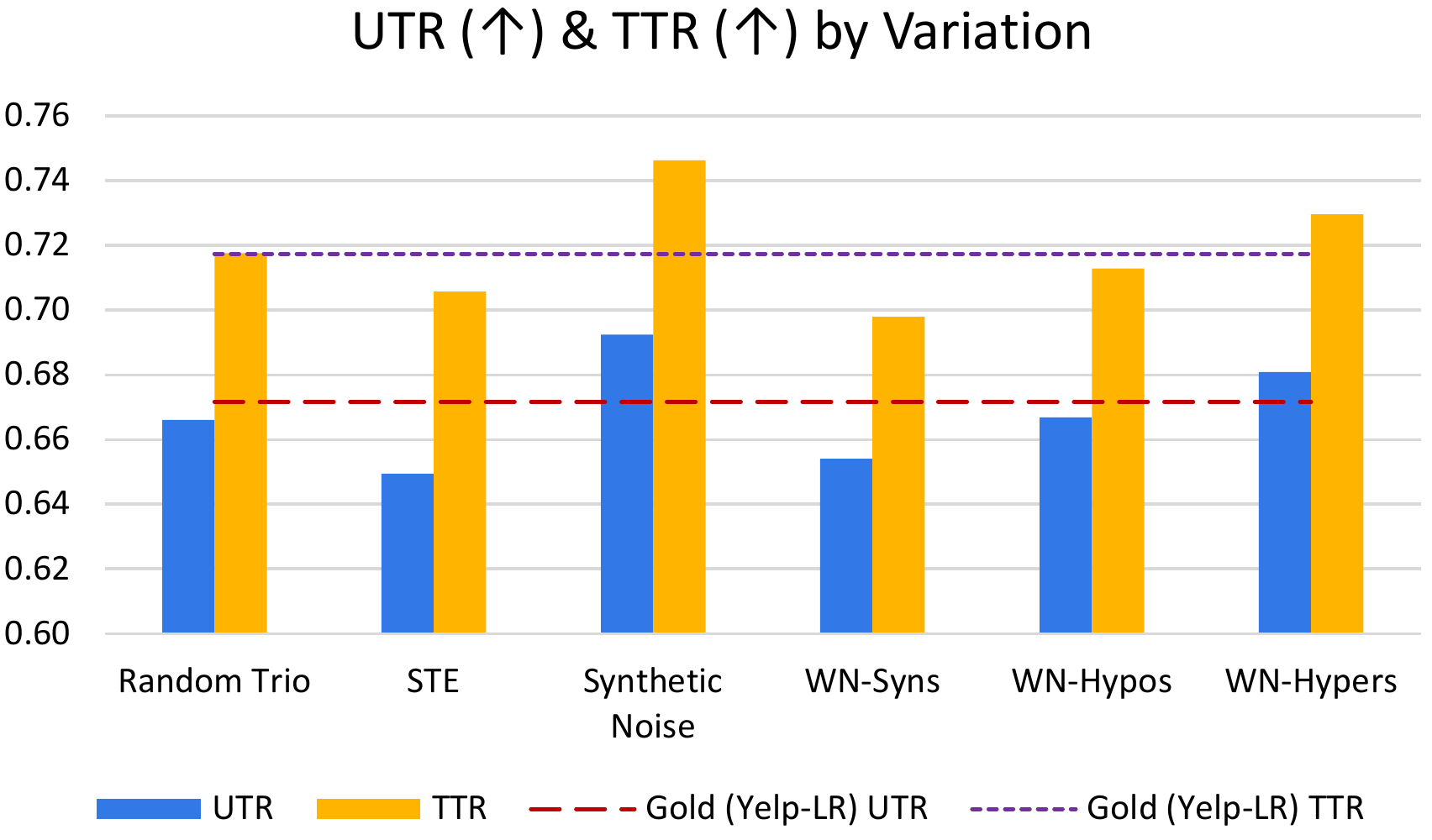}
    \caption{}
    \label{fig:variation_UTR_TTR}
\end{subfigure}
\vspace{-0.5\abovedisplayskip}
\caption{Graphs of a) average SBLEU and b) average UTR and TTR results by variation.\label{fig:overall_variation_diversity}}
\end{figure}

\begin{figure}
\begin{tabular}{@{}ll@{}}
\includegraphics[width=0.47\textwidth]{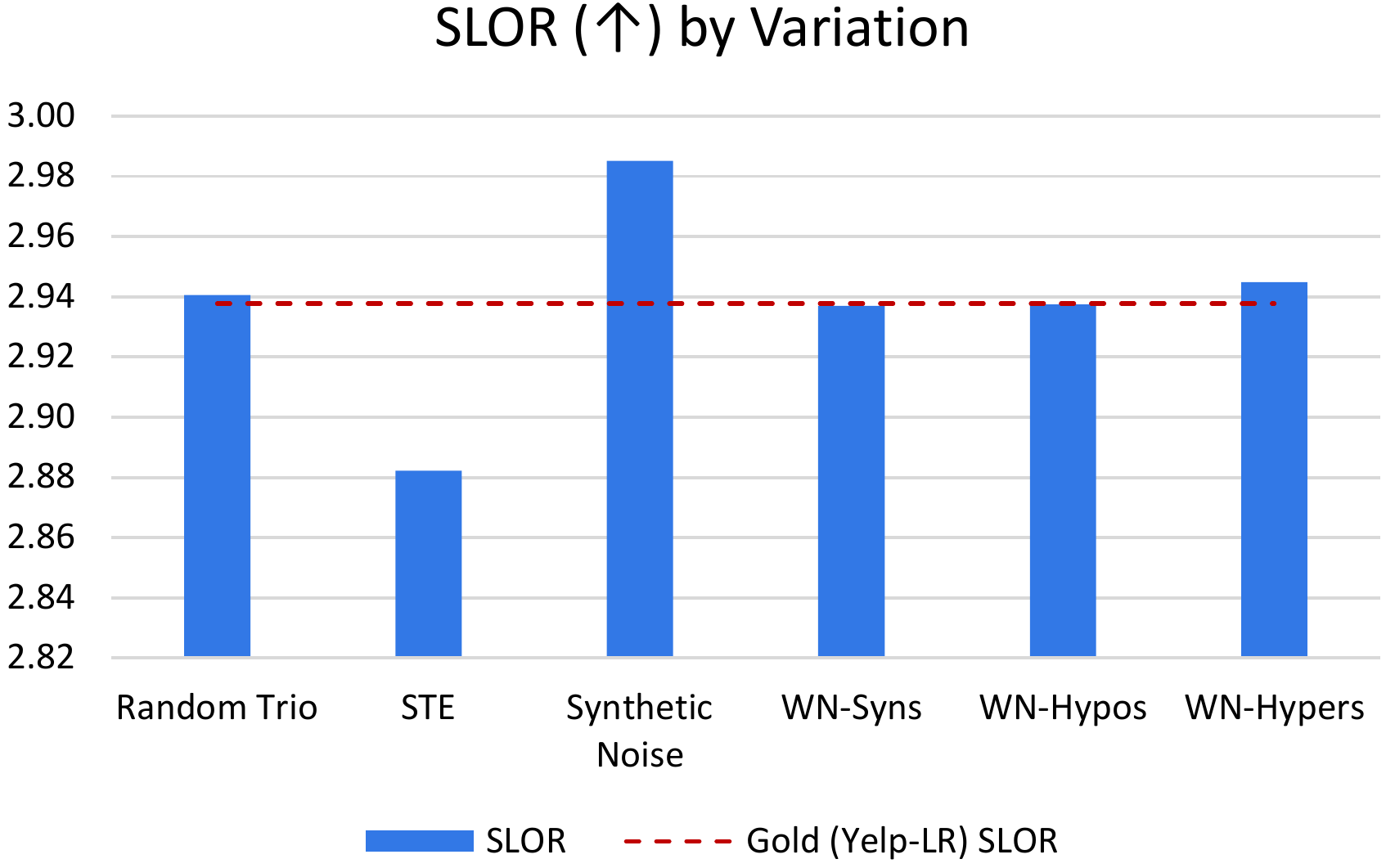} \\
\end{tabular}
  \caption{\label{fig:variation_SLOR} Graph of average SLOR results by variation.}
\end{figure}

\begin{figure}
\begin{tabular}{@{}ll@{}}
\includegraphics[width=0.47\textwidth]{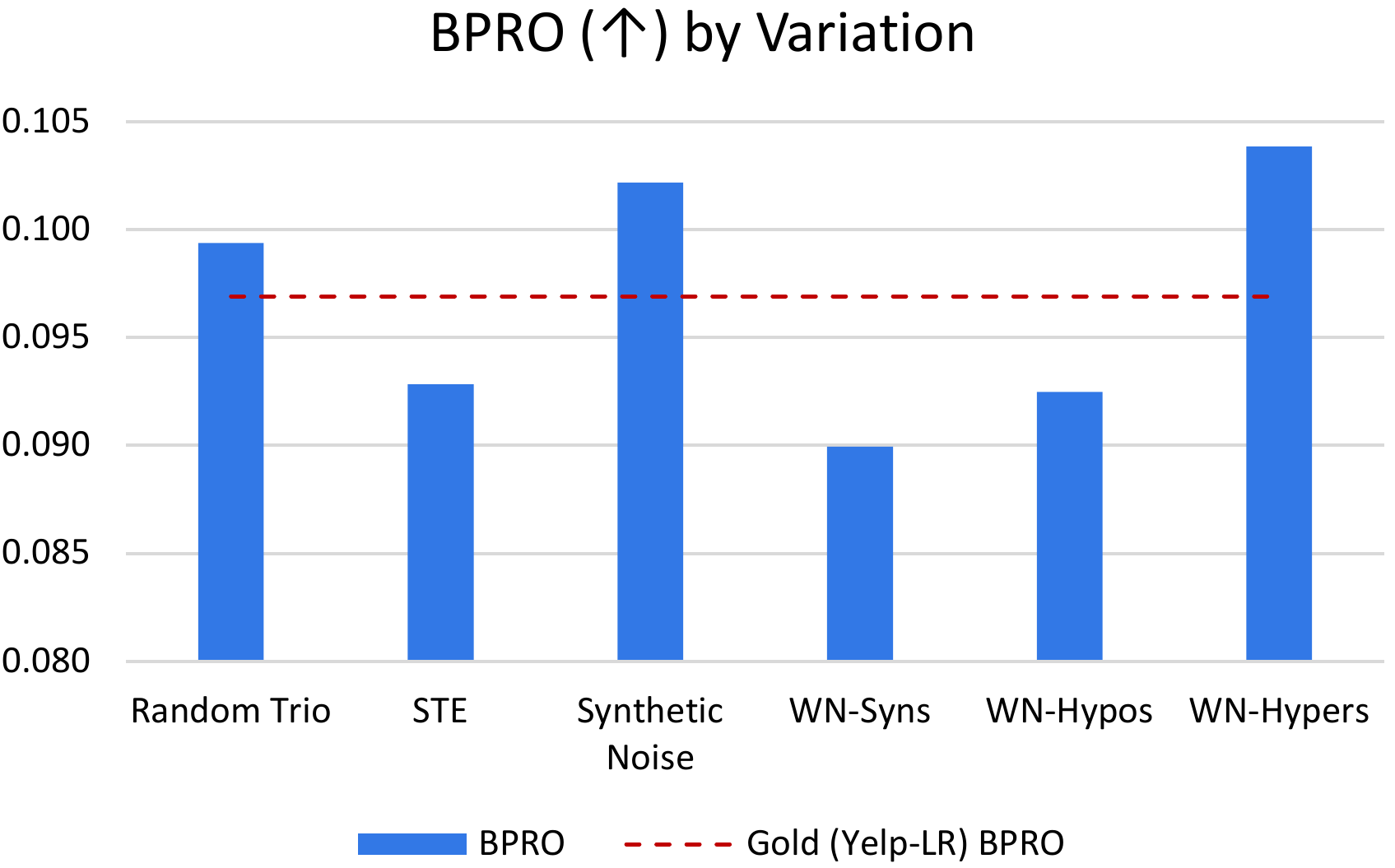} \\
\end{tabular}
  \caption{\label{fig:variation_BPRO} Graph of average BPRO results by variation.}
\end{figure}

\begin{figure}
\begin{tabular}{@{}ll@{}}
\includegraphics[width=7.7cm, height=4.83cm]{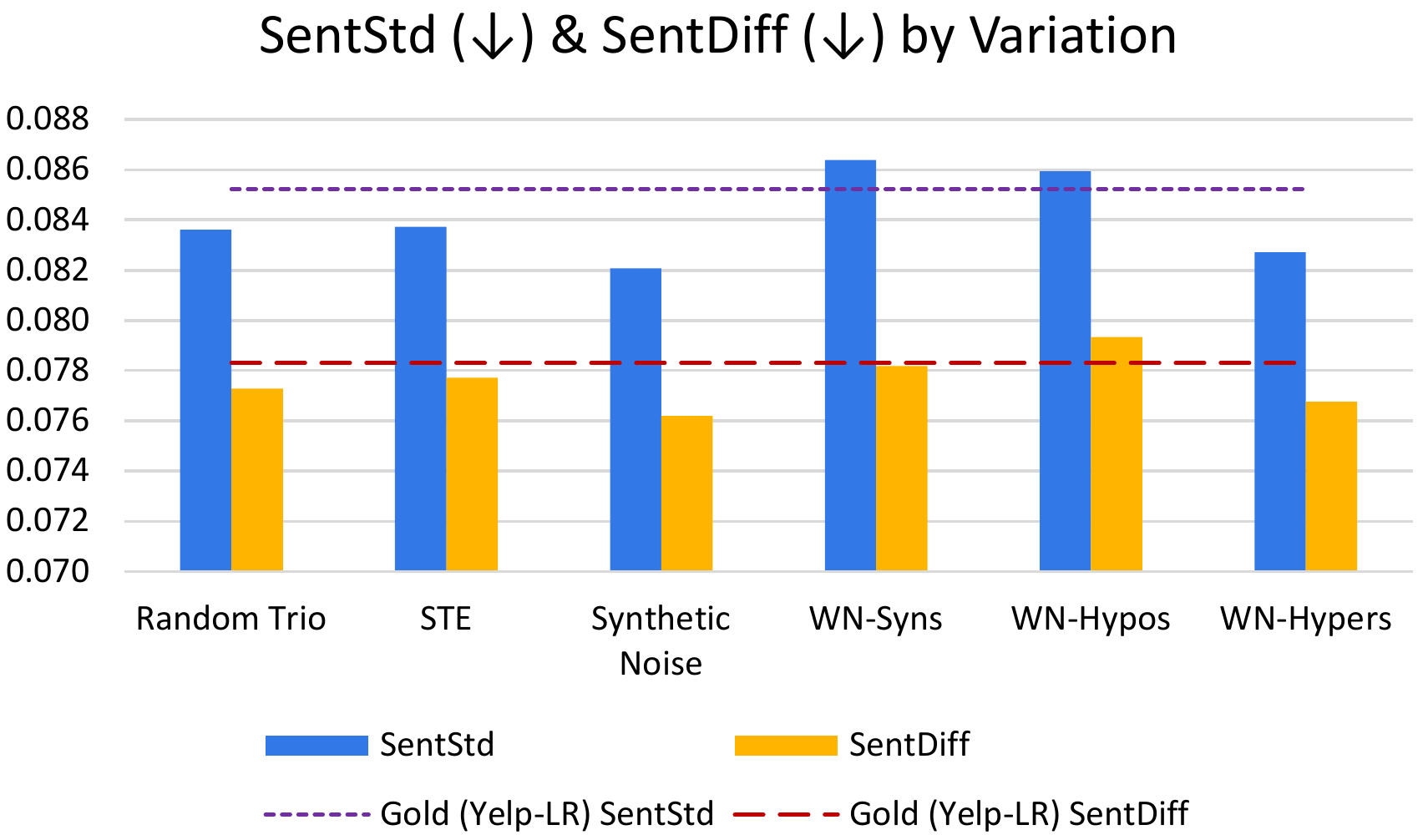} \\
\end{tabular}
  \caption{\label{fig:variation_SentStd_SentDiff} Graph of avg. sentiment results by variation.}
\end{figure}

\begin{figure}[ht]
\begin{subfigure}{.97\columnwidth}
    \centering
    \includegraphics[width=0.99\columnwidth]{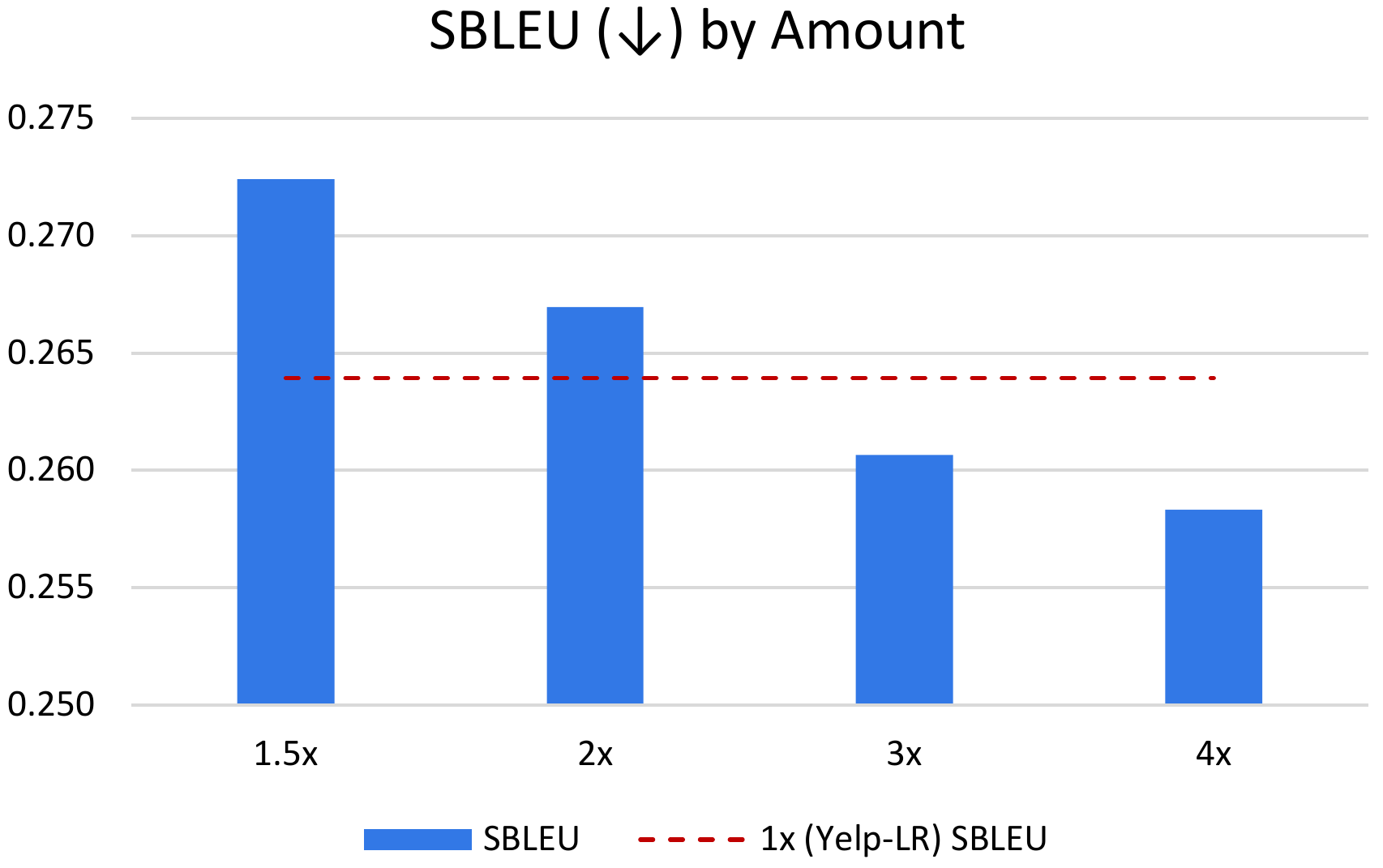}
    \caption{}
    \label{fig:amount_SBLEU}
\end{subfigure} \\
\begin{subfigure}{.97\columnwidth}
    \centering
    \includegraphics[width=0.99\columnwidth]{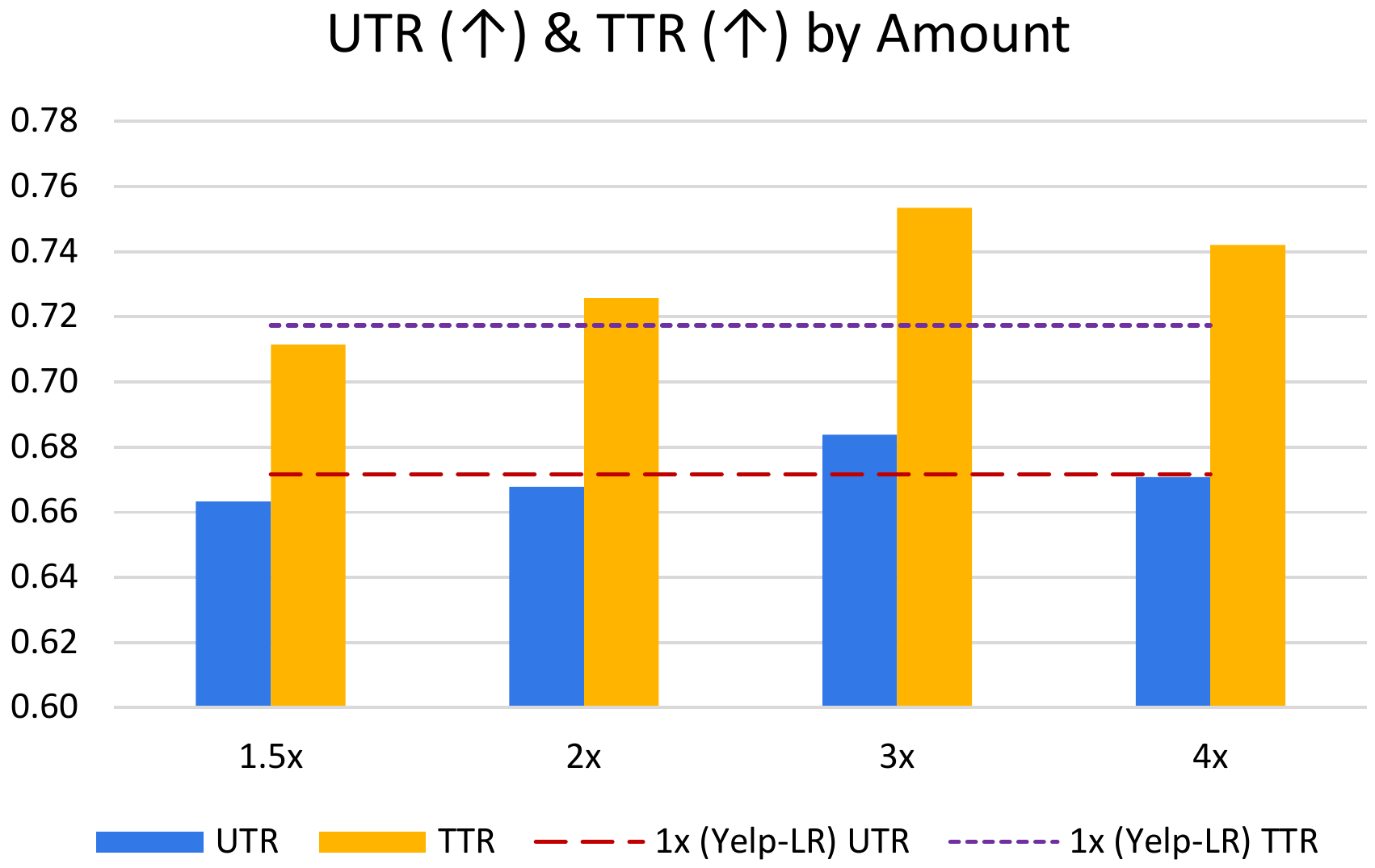}
    \caption{}
    \label{fig:amount_UTR_TTR}
\end{subfigure}
\vspace{-0.5\abovedisplayskip}
\caption{Graphs of a) average SBLEU and b) average UTR and TTR results by amount.\label{fig:overall_amount_diversity}}
\end{figure}

\begin{figure}
\begin{tabular}{@{}ll@{}}
\includegraphics[width=0.47\textwidth]{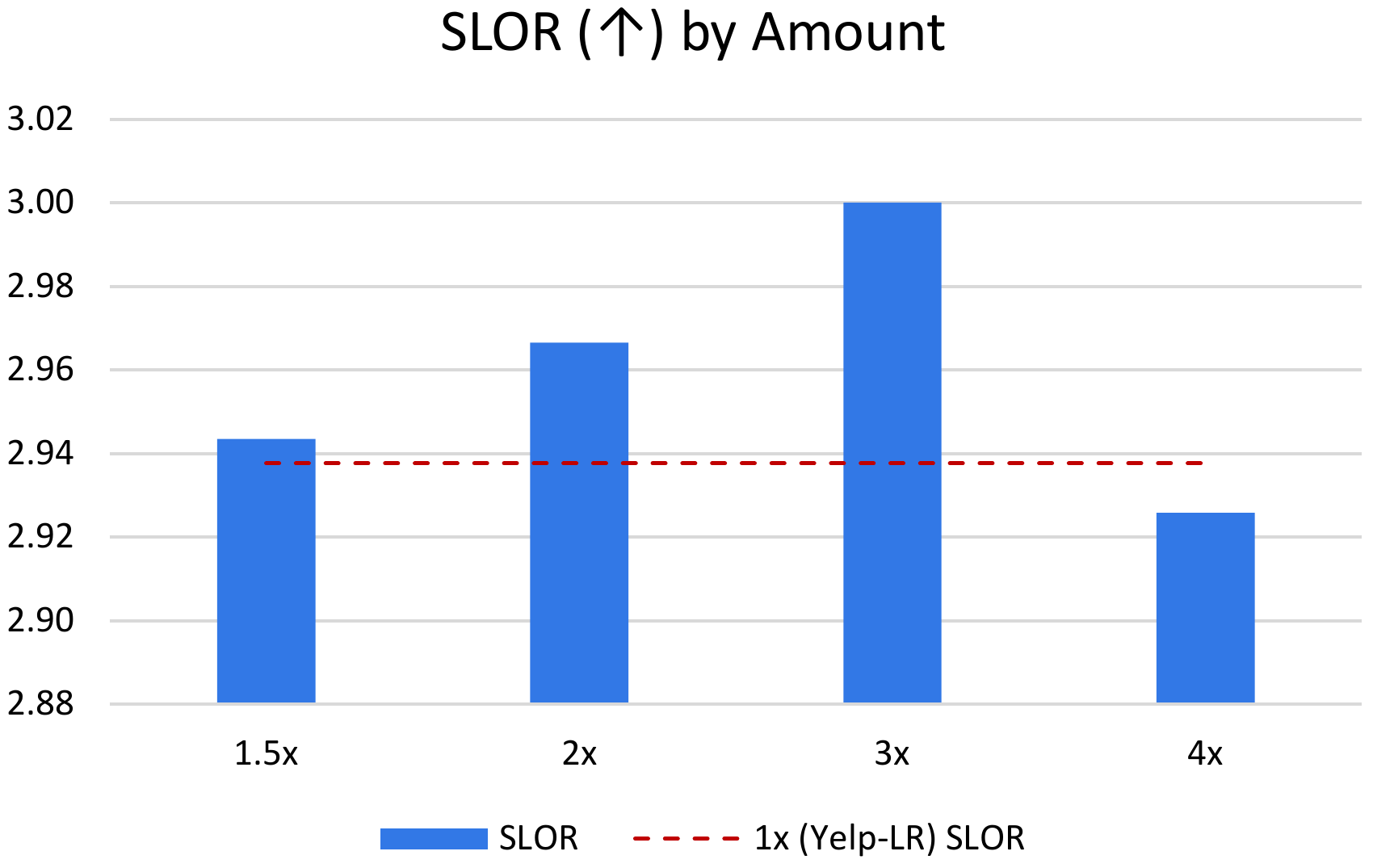} \\
\end{tabular}
  \caption{\label{fig:amount_SLOR} Graph of average SLOR results by amount.}
\end{figure}

\begin{figure}
\begin{tabular}{@{}ll@{}}
\includegraphics[width=0.47\textwidth]{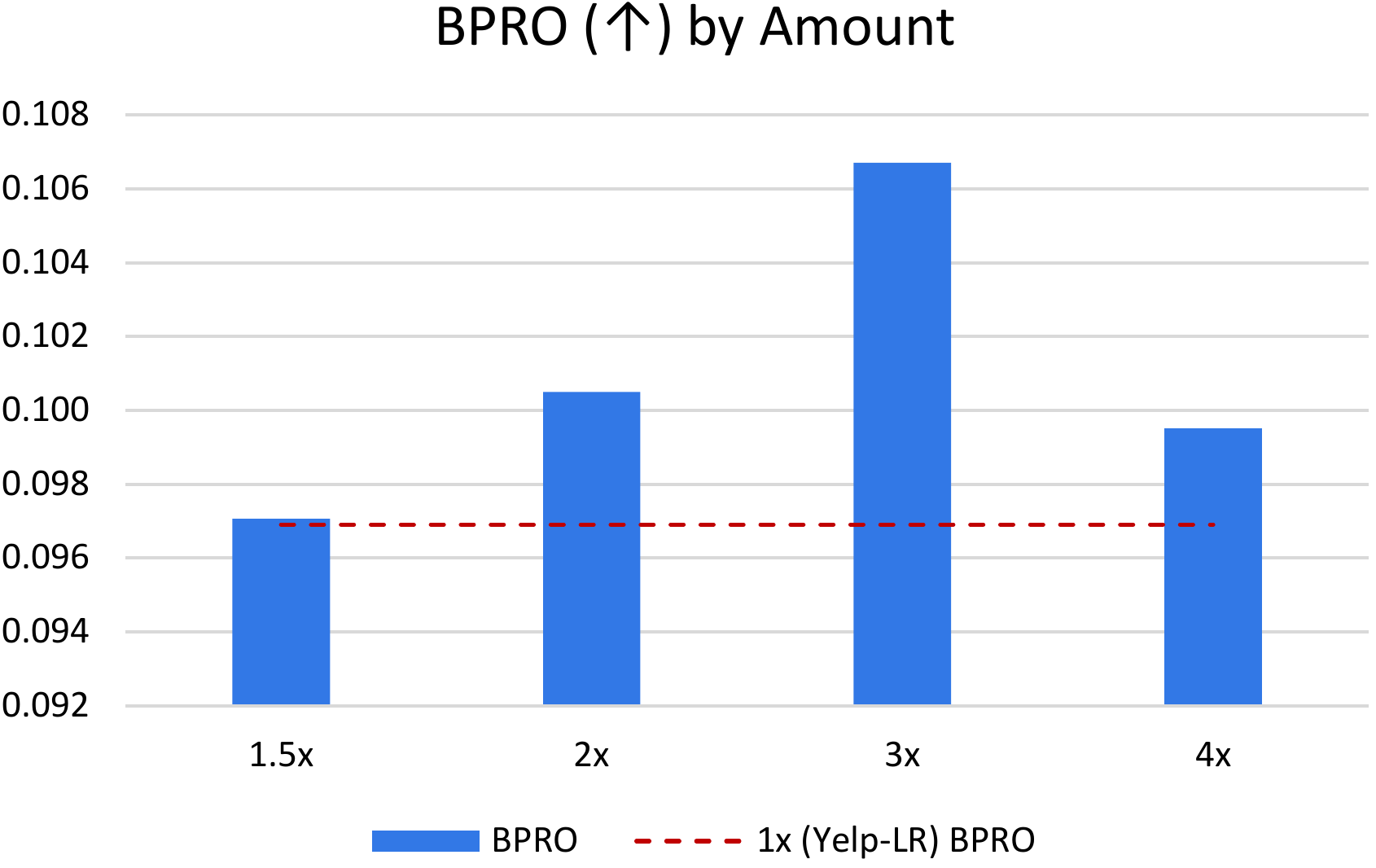} \\
\end{tabular}
  \caption{\label{fig:amount_BPRO} Graph of average BPRO results by amount.}
\end{figure}

\begin{figure}
\begin{tabular}{@{}ll@{}}
\includegraphics[width=7.5cm, height=5cm]{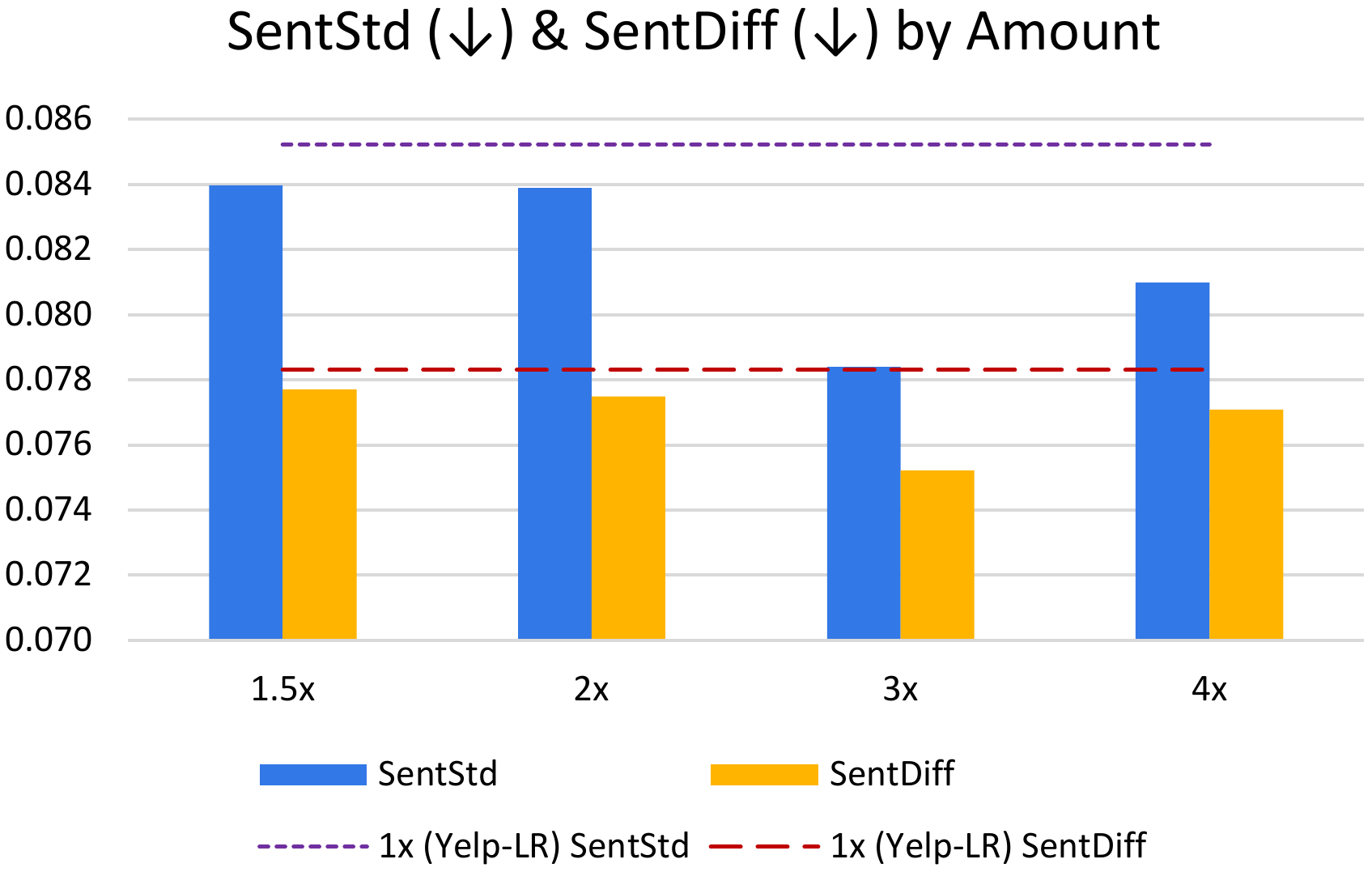} \\
\end{tabular}
  \caption{\label{fig:amount_SentStd_SentDiff} Graph of avg. sentiment results by amount.}
\end{figure}

Tables~\ref{tab:result_variations} and~\ref{tab:result_amounts} contain average evaluation results for the variations and amounts, respectively. See Appendix \S\ref{sec:appendix_significance_results} for significance p-values and Appendix \S\ref{sec:appendix_perplexity_results} for PPL results.\footnote{Statistical significances are from paired two-tailed t-tests between the Yelp-LR and particular variation and amount results using an $\alpha$ of 0.05.} Figures~\ref{fig:overall_variation_diversity} to~\ref{fig:variation_SentStd_SentDiff} contain graphs of the variation results, and Figures~\ref{fig:overall_amount_diversity} to \ref{fig:amount_SentStd_SentDiff} contain graphs of the amount results. The horizontal line(s) on the graphs refer to the no-augmentation (gold and 1x) setting with Yelp-LR. Table~\ref{tab:qualitative_2} contains generation examples.\footnote{See Appendix \S\ref{sec:appendix_continuation_examples} for more example generations.} 

\begin{table*}
\begin{tabular}{@{}ll@{}}
\includegraphics[width=0.98\textwidth]{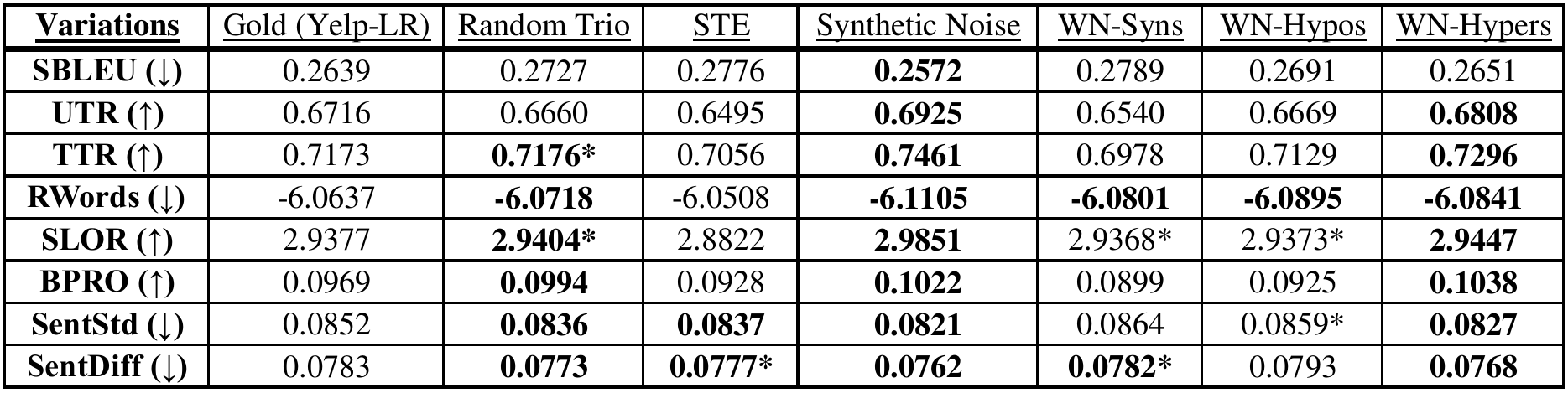}\\
\end{tabular}
  \vspace{-0.5\abovedisplayskip}
  \caption{\label{tab:result_variations} Average results by variation. Bold values indicate results better than Gold (Yelp-LR). Arrows beside each metric indicate whether lower or higher is better. * indicates insignificant values (using an $\alpha$ of 0.05).}
\end{table*}

\begin{table}
\centering
\begin{tabular}{@{}ll@{}}
\includegraphics[width=0.46\textwidth]{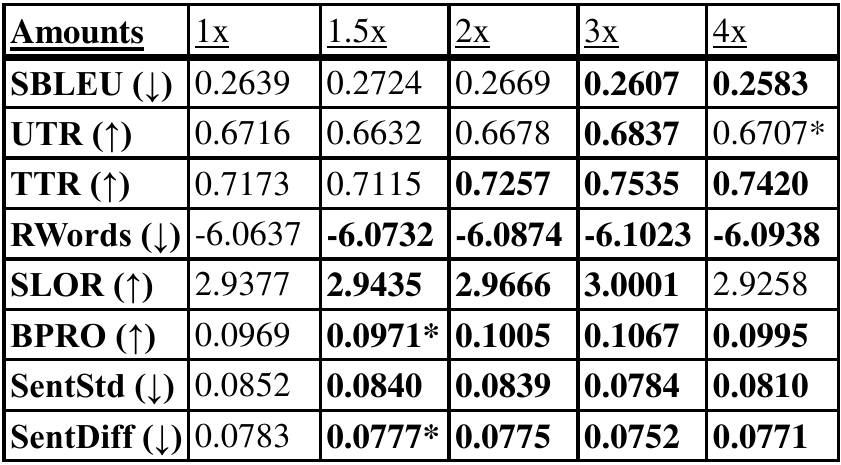}\\
\end{tabular}
  \vspace{-0.5\abovedisplayskip}
  \caption{\label{tab:result_amounts} Average results by amount. Bold values indicate results better than 1x (Yelp-LR). Arrows beside each metric indicate whether lower or higher is better. * indicates insignificant values (using an $\alpha$ of 0.05).}
\end{table}



\begin{table*}
\begin{tabular}{ccc}
\includegraphics[width=0.95\textwidth]{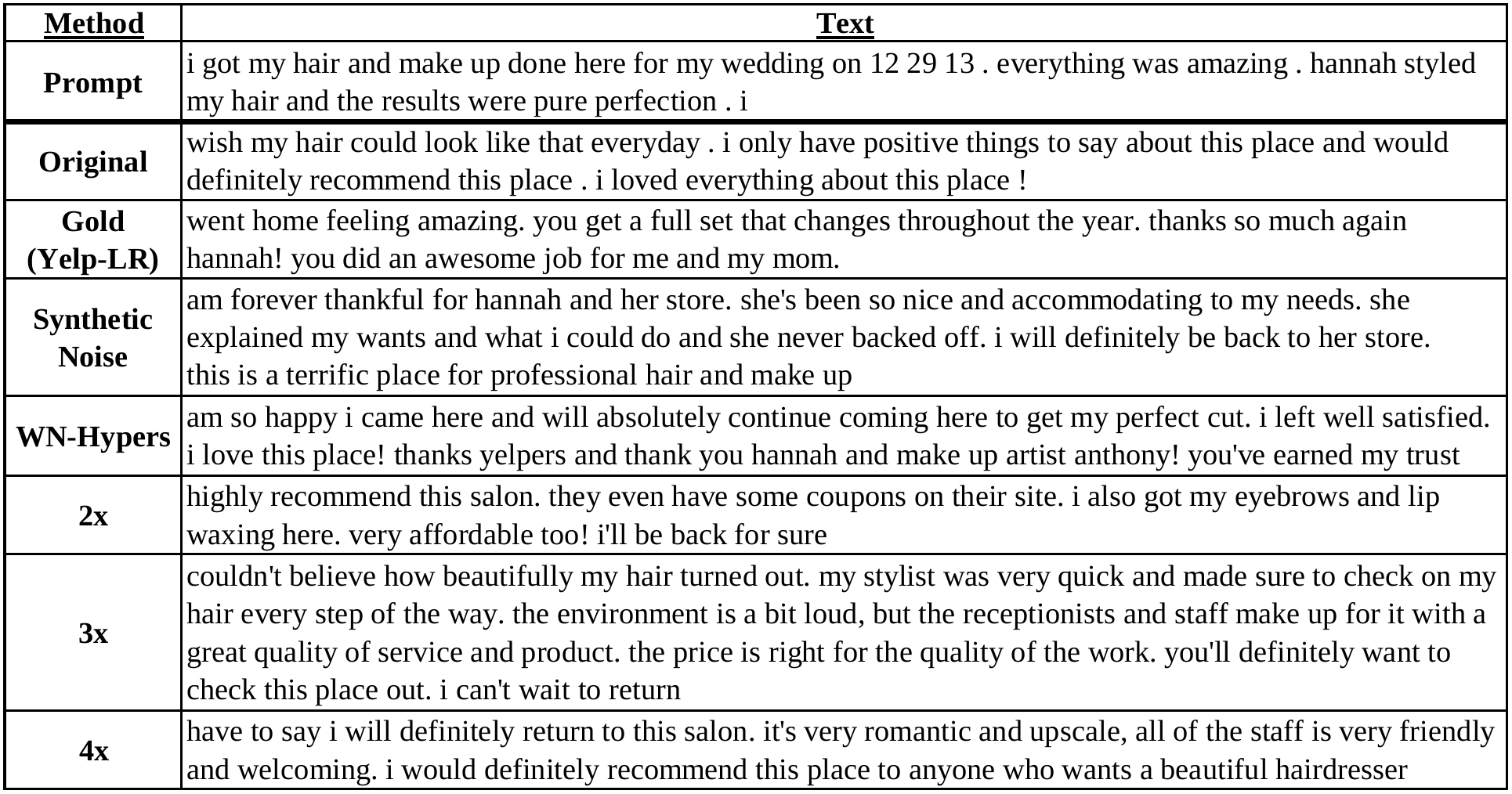}\\
\end{tabular}
  \vspace{-0.5\abovedisplayskip}
  \caption{\label{tab:qualitative_2} Examples of generated continuations from GPT-2 finetuned on select augmentation methods \& amounts. \textit{Prompt} is the first half of the original Yelp review fed in as input, and \textit{Original} is the ground-truth continuation.}
\end{table*}

\subsection{Performance by Augmentation Method}
\label{sec:performance_by_method}
We analyze the performance of each augmentation method using Table \ref{tab:result_variations} and Figures~\ref{fig:overall_variation_diversity} to \ref{fig:variation_SentStd_SentDiff}.

\subsubsection{Synthetic Noise and WN-Hypers}
Synthetic Noise beats gold considerably on every metric. WN-Hypers does as well (other than SBLEU), but to a lesser extent on most metrics. Both clearly improve upon the gold setting.

To ensure Synthetic Noise's diversity improvements are not due to increased misspellings, we measure SpellWords and SpellChars. As seen in Table~\ref{tab:result_spellcheck}, Synthetic Noise actually decreases the average number of misspellings. This is likely because we only insert noise into the prompt portion of the training reviews, and GPT-2 is learning to be more robust to noise when finetuned on this data. This may also lead to increased generation quality. 

WN-Hypers may improve performance as it slightly semantically adjusts the text. It does not keep semantics the same (unlike the goal of WN-Syns) but also does not drift too far since we choose the closest hypernyms. Each one carries more high-level meaning, which may contribute to increasing text diversity and fluency. We hence show that the integration of external knowledge for GenAug can improve performance.

Unlike WN-Hypos, where replacements can be esoteric and rare, WN-Hypers' are typically more common words with higher chance of being seen by GPT-2 while training and appearing naturally at test-time. An example is replacing \textit{dog} with \textit{animal} (WN-Hypers) vs. \textit{corgi} (WN-Hypos). Further, except quantified statements (e.g. \textit{``All dogs bark"}), most WN-Hypers examples retain \textit{faithfulness}\footnote{Sentence Y being \emph{Faithful} to Sentence X implies Y does not \emph{hallucinate} or state information not already implied by X.} \cite{maynez2020faithfulness} to the original, e.g. \textit{``3 dogs walked home"} entails \textit{``3 animals walked home"}.

\subsubsection{STE and WN-Syns}
STE and WN-Syns perform noticeably worse than gold. STE decreases fluency, diversity, and BPRO, albeit the sentiment-related metrics improve. WN-Syns decreases diversity and BPRO.

A possible explanation for STE is that SMERTI works best for shorter text.\footnote{See Section \S\ref{subsection:STE} for an explanation.} Our sliding window is also problematic as text between windows may have semantic inconsistencies. For example, in Table~\ref{tab:qualitative_1}, the chosen REs are \textit{coffee} and \textit{hand}; \textit{hand} results in \textit{washing machine}, making the last part semantically inconsistent with the first part about \textit{coffee}. This likely results in reduced fluency and BPRO. Reduced fluency is also not unexpected as \citet{feng2019keep} showed STE reduces SLOR.

A possible explanation for WN-Syns is that synonyms keep the semantic content almost exactly the same, unlike the other augmentation techniques which vary the semantics of the text more. Hence, GPT-2 may be overfitting to the augmented data.

\begin{table}
\centering
\begin{tabular}{ccc}
\includegraphics[width=0.40\textwidth]{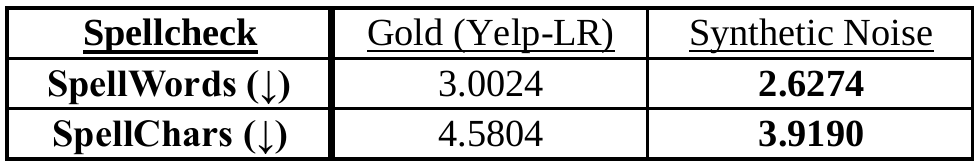}\\
\end{tabular}
  \vspace{-0.7\abovedisplayskip}
  \caption{\label{tab:result_spellcheck} Average Spellcheck results.}
\end{table}

\subsubsection{WN-Hypos and Random Trio}
Lastly, WN-Hypos and Random Trio also do not perform very well. WN-Hypos performs worse than gold on almost all metrics, but to a lesser extent. For Random Trio, overall diversity is decreased, but BPRO and sentiment-related metrics improve. A couple of Random Trio's metric improvements are minor and statistically insignificant. 

This is likely due to Random Trio's techniques involving almost complete randomness (at the word-level), resulting in high variations in the metric results, leading to statistical insignificance and poor generations. Its random techniques appear much less suitable for GenAug than data augmentation for classification \cite{wei2019eda}. 

\phantomsection
\label{para:hyponyms}
For WN-Hypos, we observe that some hyponyms diverge more from the parent than others (e.g. \textit{food} $\rightarrow$ \textit{beverage} vs. \textit{food} $\rightarrow$ \textit{micronutrient}), which can cause large drifts in meaning. Similar to Random Trio, this word-level \textit{randomness} is likely leading to poor generations. Further, many hyponyms are esoteric words that GPT-2 has likely rarely (or never) seen (e.g \textit{dragon}$\rightarrow$\textit{wyvern}), further decreasing performance. See the example in Table~\ref{tab:qualitative_1} (notice the word \textit{kiribati}). Hence, we show that incorporation of external knowledge for GenAug can also decrease performance.

\subsubsection{Overall Performance}
Overall, Synthetic Noise and WN-Hypernyms are the best performing methods for GenAug on YLR (see Table~\ref{tab:qualitative_2} for example generations), and the others perform noticeably worse and are hence not recommended in their current state.

\subsection{Performance by Augmentation Amount}
Table \ref{tab:result_amounts} and Figures~\ref{fig:overall_amount_diversity} to \ref{fig:amount_SentStd_SentDiff} show that quality of the generated text improves from 1.5x to 3x data augmentation, and decreases from 3x to 4x (except for SBLEU). 3x beats gold considerably on every metric, while 2x and 4x beat gold noticeably on most metrics as well (see Table~\ref{tab:qualitative_2} for example continuations). 1.5x performs noticeably worse than gold on text diversity. 

Quality of the text really improves from 2x and onward, reaching a peak at 3x, and dropping afterward (especially in SLOR). For GenAug on YLR, 3x augmentation appears optimal, and more can reduce performance. This could be attributed to overfitting since many augmentation methods modify the original text to a limited degree. Augmentation at high amounts would thus have a similar (but lesser) effect to training on repeated examples.


%% file: sections/related-work.tex
There has been work using GPT-2 as a component in the data augmentation process for training classifiers \cite{kumar2020data,papanikolaou2020dare}. We investigate augmentation for \emph{finetuning GPT-2 itself}, and in fact deal with a precondition for the former - without a language model conforming to the domain, generated text would be further from the domain distribution.

There is also work on data augmentation for training NLP classifiers such as \citet{wei2019eda}, \citet{lu2006}, and \citet{kobayashi-2018-contextual}. We adopt some techniques from \citet{wei2019eda} for our experiments, but in general, augmentation techniques for classification do not necessarily work well for generation. The distribution learned in the latter case, $P(x_{c}|x), x_{c} \in {|V|}^*$, is more complex than the former, $P(y|x), y \in Y\subset N$, due to a higher dimensional output variable (where $Y$ is the label set, $x_{c}$ denotes continuation, and $|V|$ refers to the vocabulary).

Generation of adversarial examples (AVEs) to evaluate robustness of NLP tasks is another area being investigated. \citet{jia2017adversarial} construct AVEs for span-based QA by adding sentences with distractor spans to passages. \citet{zhang2019paws} use word swapping to craft AVEs for paraphrase detection. Unlike these works, we are not concerned with test-time invariance or test-time model behavior on augmented examples, as long as these augmented examples improve training.

\citet{kang2018adventure} and \citet{glockner2018breaking} use WordNet relations to construct AVEs for textual entailment. However, to the best of our knowledge, we are the first ones to explore such methods using WordNet and lexical databases for text data augmentation for generative models.

%% file: sections/conclusions.tex
We introduced and investigated \textit{GenAug}: data augmentation for text generation, specifically finetuning text generators, through various augmentation methods. We finetuned GPT-2 on a subset of the Yelp Reviews dataset, and demonstrated that insertion of character-level synthetic noise and keyword replacement with hypernyms are effective augmentation methods. We also showed that the quality of generated text improves to a peak at approximately three times the amount of original training data.

Potential future directions include exploring augmentation based on a) linguistic principles like compositionality \cite{andreas2019good} and b) using more complex lexical resources - e.g. Framenet \cite{baker1998berkeley}. One can also investigate further augmentation techniques using word replacement such as exploring the \textit{contextual augmentation} method used in \citet{kobayashi-2018-contextual}. Further, methods of improving semantic text exchange (STE) on longer texts can be investigated, which would make it more effective for data augmentation. Lastly, there is potential in exploring data augmentation for other domains such as dialogue and related tasks such as style transfer \cite{kang-etal-2019-male}, and investigating interesting aspects of it such as dialogue personalization \cite{li2020aloha}.

%% file: sections/appendix.tex
\begin{table*}
\begin{tabular}{@{}ll@{}}
\includegraphics[width=0.98\textwidth]{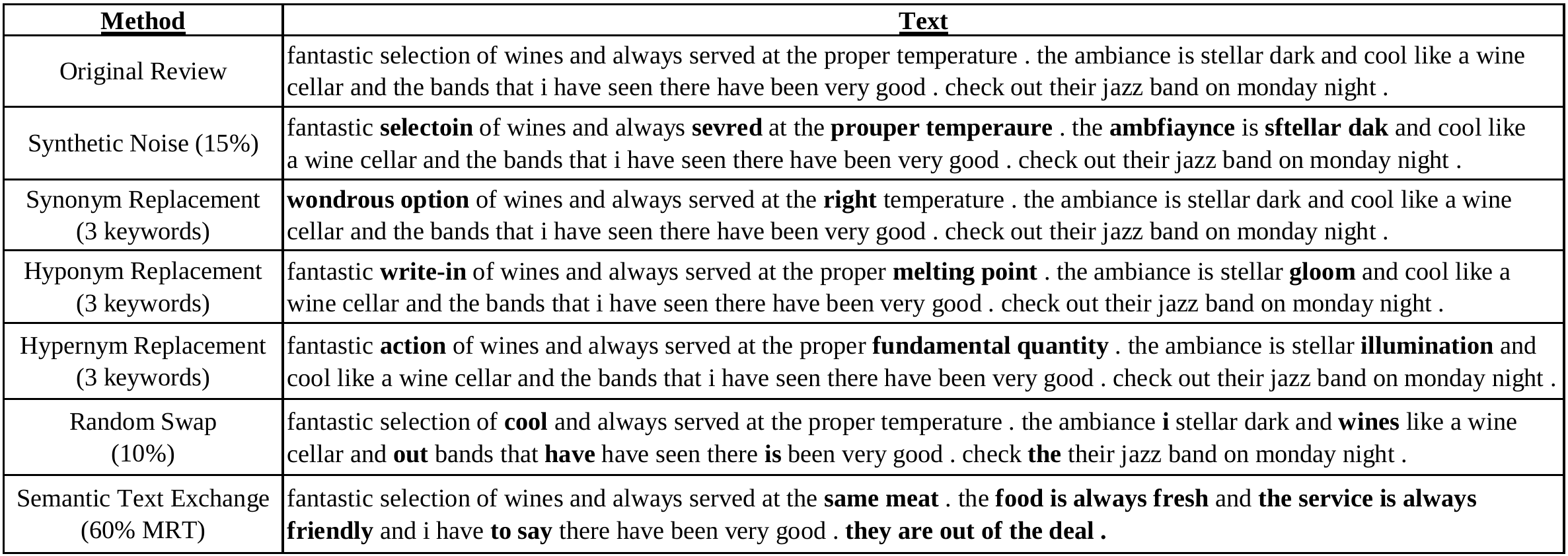}\\
\end{tabular}
  \caption{\label{tab:qualitative_appendix_1} Example of a Yelp review and its variations using our augmentation methods. Changes are bolded.}
\end{table*}

\begin{table*}
\begin{tabular}{@{}ll@{}}
\includegraphics[width=0.98\textwidth]{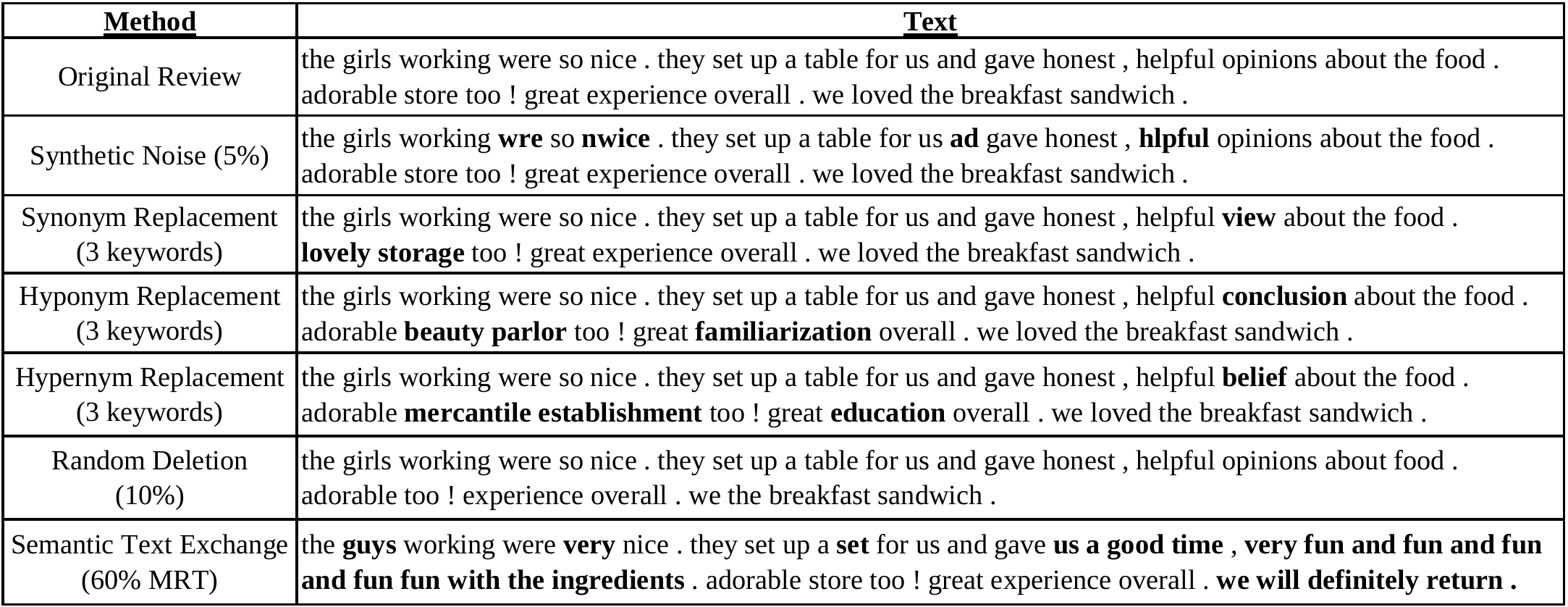}\\
\end{tabular}
  \caption{\label{tab:qualitative_appendix_2} Example of a Yelp review and its variations using our augmentation methods. Changes are bolded (except for Random Deletion where words were removed).}
\end{table*}

\begin{table*}
\begin{tabular}{ccc}
\includegraphics[width=0.95\textwidth]{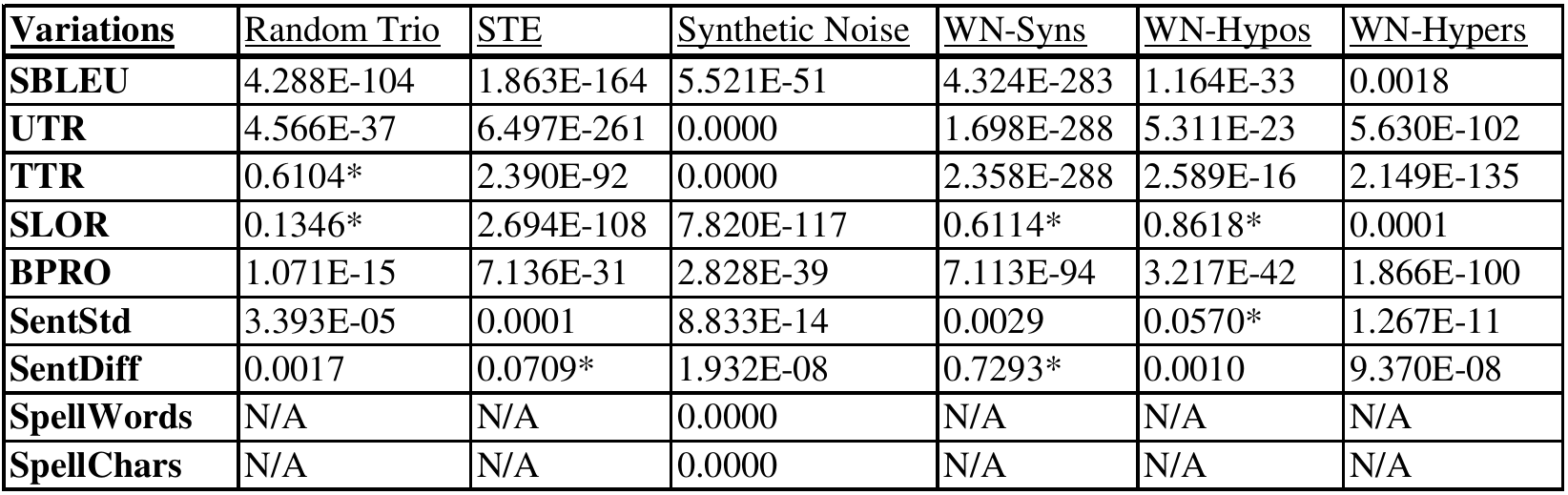}\\
\end{tabular}
  \caption{\label{tab:sig_variations} p-values of results by variation. Note: * indicates insignificant values (using an alpha of 0.05).}
\end{table*}

\begin{table*}
\begin{tabular}{ccc}
\includegraphics[width=0.95\textwidth]{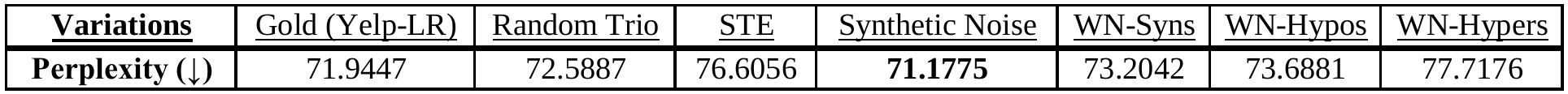}\\
\end{tabular}
  \caption{\label{tab:ppl_variations} Average perplexity results by variation. Note: bold values are better (lower) than gold (Yelp-LR).}
\end{table*}

\begin{table*}
\centering
\begin{tabular}{ccc}
\includegraphics[width=0.70\textwidth]{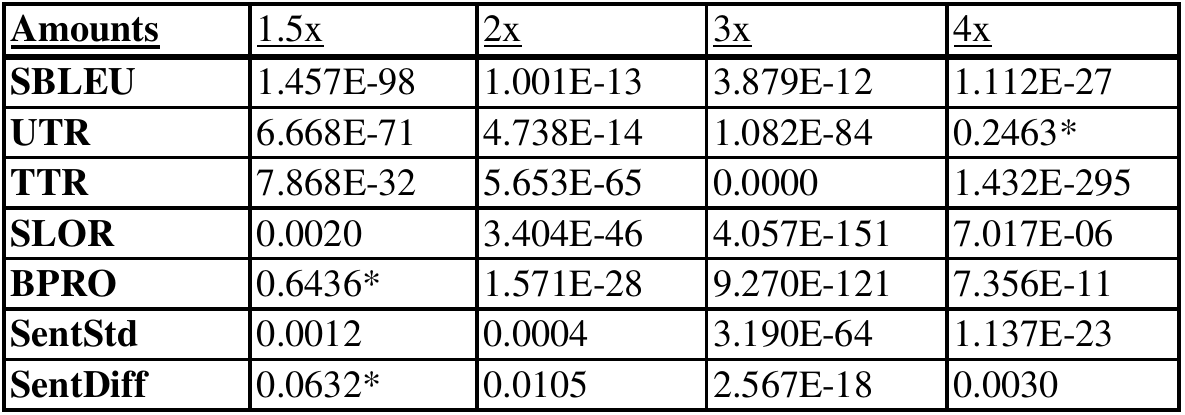}\\
\end{tabular}
  \caption{\label{tab:sig_amounts} p-values of results by amount. Note: * indicates insignificant values (using an alpha of 0.05).}
\end{table*}
\begin{table*}
\centering
\begin{tabular}{ccc}
\includegraphics[width=0.70\textwidth]{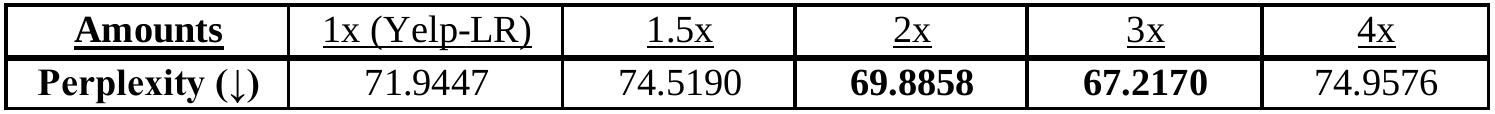}\\
\end{tabular}
  \caption{\label{tab:ppl_amounts} Average perplexity results by amount. Note: bold values are better (lower) than 1x (Yelp-LR).}
\end{table*}

\begin{table*}[ht]
\begin{tabular}{@{}ll@{}}
\includegraphics[width=0.98\textwidth]{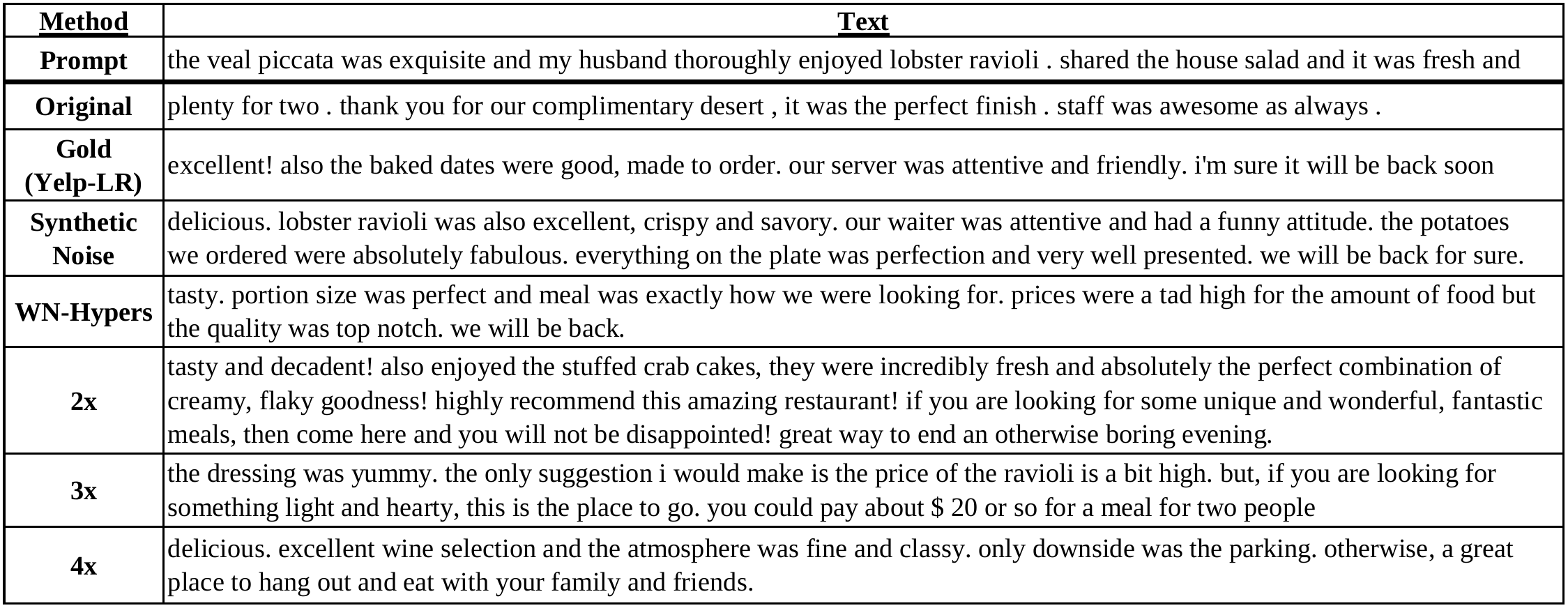}\\
\end{tabular}
  \vspace{-0.5\abovedisplayskip}
  \caption{\label{tab:qualitative_appendix_3} Examples of generated continuations from GPT-2 finetuned on select augmentation methods \& amounts. \textit{Prompt} is the first half of the original Yelp review fed in as input, and \textit{Original} is the ground-truth continuation.}
\vspace{0.6cm} 
\begin{tabular}{@{}ll@{}}
\includegraphics[width=0.98\textwidth]{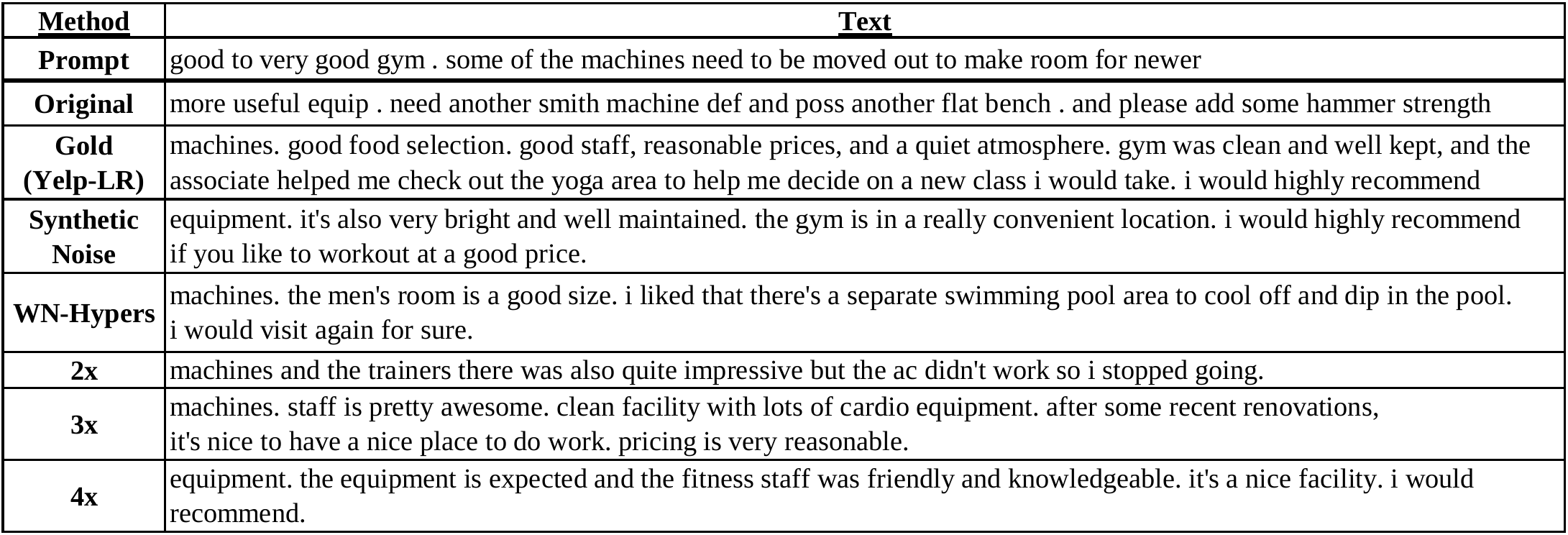}\\
\end{tabular}
  \vspace{-0.5\abovedisplayskip}
  \caption{\label{tab:qualitative_appendix_4} Examples of generated continuations from GPT-2 finetuned on select augmentation methods \& amounts. \textit{Prompt} is the first half of the original Yelp review fed in as input, and \textit{Original} is the ground-truth continuation.}
\end{table*}

\section{Augmentation Variation Examples}
\label{sec:appendix_augmentation_variation_examples}

See Tables~\ref{tab:qualitative_appendix_1} and \ref{tab:qualitative_appendix_2} for further examples of Yelp review variations using our augmentation methods.

\section{SMERTI Sliding Window Algorithm}
\label{sec:appendixWindowDetails}
We use 30-word windows, consisting of 10 words of context (the last 10 words of the previous window) and 20 new words.\footnote{The first window is 20 words long and has no context. If a review is at most 25 words long, we perform STE on the entire review (without the sliding window algorithm).} In the context portion of each window, we cannot insert the RE nor mask or replace any words. In the new 20-word portion of each window, we can insert the new RE and mask and replace other words. This ensures when SMERTI performs STE on each window, it is able to utilize some context from the previous window but is unable to modify and blemish the STE already performed on the previous window.

\section{Sentiment Regressor Finetuning}
\label{sec:appendix_sentiment_regressor_finetune}
The BERT sentiment regressor is finetuned on the same Yelp-LR 50K training and 15K validation splits. The final classifer we use is after three epochs of finetuning. Details as follows:
\begin{itemize}[noitemsep,topsep=2pt]
  \item Star rating conversion: 1 star = 0, 2 star = 0.25, 3 star = 0.5, 4 star = 0.75, 5 star = 1
  \item Finetuning details:
  \begin{itemize}[noitemsep,topsep=0pt]
        \item max\_seq\_length: 128
        \item per\_gpu\_eval\_batch\_size: 32
        \item per\_gpu\_train\_batch\_size: 32
        \item learning\_rate: 2e-5
  \end{itemize}
\end{itemize}

\section{Finetuned Model Details}
\label{subsec:finetuned_model_details}
Note: BE below stands for ``best epoch", and VPPL for ``validation perplexity".
\begin{itemize}[noitemsep,topsep=2pt]
    \item Two-million review subset of Yelp (for PPL and SLOR eval): BE = 4, VPPL = 9.1588
    \item Seed set 1 finetuned models:
    \begin{itemize}[noitemsep,topsep=0pt]
        \item gpt2\_gold: BE = 3, VPPL = 11.7309
        \item gpt2\_noise: BE = 3, VPPL = 12.0408
        \item gpt2\_STE: BE = 3, VPPL = 12.1892
        \item gpt2\_syns: BE = 2, VPPL = 11.9844
        \item gpt2\_hypos: BE = 2, VPPL = 11.9638
        \item gpt2\_hypers: BE = 2, VPPL = 12.0131
        \item gpt2\_random: BE = 2, VPPL = 11.9297
        \item gpt2\_1.5x: BE = 3, VPPL = 11.8958
        \item gpt2\_2x: BE = 3, VPPL = 11.9113
        \item gpt2\_3x: BE = 2, VPPL = 12.2064
        \item gpt2\_4x: BE = 1, VPPL = 12.3574
    \end{itemize}
    \item Seed set 2 finetuned models:
    \begin{itemize}[noitemsep,topsep=0pt]
        \item gpt2\_gold: BE = 3, VPPL = 11.7387
        \item gpt2\_noise: BE = 2, VPPL = 12.0230
        \item gpt2\_STE: BE = 3, VPPL = 12.1711
        \item gpt2\_syns: BE = 2, VPPL = 11.9282
        \item gpt2\_hypos: BE = 2, VPPL = 11.9583
        \item gpt2\_hypers: BE = 2, VPPL = 11.9957
        \item gpt2\_random: BE = 2, VPPL = 11.9558
        \item gpt2\_1.5x: BE = 3, VPPL = 11.8943
        \item gpt2\_2x: BE = 2, VPPL = 12.0209
        \item gpt2\_3x: BE = 2, VPPL = 12.1710
        \item gpt2\_4x: BE = 1, VPPL = 12.3288
    \end{itemize}
\end{itemize}

\section{SMERTI-Transformer Training}
\label{sec:appendixSMERTITTDetails}
Similar to~\citet{feng2019keep}, we use scaled dot-product attention and the same hyperparameters as~\citet{vaswani2017attention}. We use the Adam optimizer \cite{kingma2014adam} with $\beta_1 = 0.9, \beta_2 = 0.98$, and $\epsilon = 10^{-9}$. We increase the learning rate (LR) linearly for the first $warmup\_steps$ training steps, and then decrease the LR proportionally to the inverse square root of the step number. We set $factor=1$, $warmup\_steps = 2000$, and use a batch size of 4096.

\section{Statistical Significance p-values}
\label{sec:appendix_significance_results}
See Tables~\ref{tab:sig_variations} and~\ref{tab:sig_amounts} for p-values of results by variation and amount, respectively. These are the results from paired two-tailed t-tests against Yelp-LR (Gold and 1x) results. We test statistical significance of all metrics other than RWords and PPL, and use an alpha of 0.05.

\section{Perplexity (PPL) Results}
\label{sec:appendix_perplexity_results}
See Tables~\ref{tab:ppl_variations} and~\ref{tab:ppl_amounts} for average PPL results by variation and amount, respectively. Synthetic Noise, 2x, and 3x beat gold (Yelp-LR), similar to SLOR. However, WN-Hypers has higher PPL than gold (unlike SLOR). This is likely due to WN-Hypers having outputs that contain rarer tokens, thus increasing PPL. We note again that SLOR normalizes for this and is a better measure of fluency overall.

\section{Generated Continuation Examples}
\label{sec:appendix_continuation_examples}

See Tables~\ref{tab:qualitative_appendix_3} and \ref{tab:qualitative_appendix_4} for further examples of generated continuations from the various experiments.

%% file: emnlp2020.bbl
\begin{thebibliography}{37}
\expandafter\ifx\csname natexlab\endcsname\relax\def\natexlab#1{#1}\fi

\bibitem[{Andreas(2020)}]{andreas2019good}
Jacob Andreas. 2020.
\newblock \href {https://doi.org/10.18653/v1/2020.acl-main.676} {Good-enough
  compositional data augmentation}.
\newblock In \emph{Proceedings of the 58th Annual Meeting of the Association
  for Computational Linguistics}, pages 7556--7566, Online. Association for
  Computational Linguistics.

\bibitem[{Baker et~al.(1998)Baker, Fillmore, and Lowe}]{baker1998berkeley}
Collin~F Baker, Charles~J Fillmore, and John~B Lowe. 1998.
\newblock The berkeley framenet project.
\newblock In \emph{36th Annual Meeting of the Association for Computational
  Linguistics and 17th International Conference on Computational Linguistics,
  Volume 1}, pages 86--90.

\bibitem[{Belinkov and Bisk(2017)}]{belinkov2017synthetic}
Yonatan Belinkov and Yonatan Bisk. 2017.
\newblock Synthetic and natural noise both break neural machine translation.
\newblock \emph{arXiv preprint arXiv:1711.02173}.

\bibitem[{Devlin et~al.(2019)Devlin, Chang, Lee, and
  Toutanova}]{devlin2018bert}
Jacob Devlin, Ming-Wei Chang, Kenton Lee, and Kristina Toutanova. 2019.
\newblock \href {https://doi.org/10.18653/v1/N19-1423} {{BERT}: Pre-training of
  deep bidirectional transformers for language understanding}.
\newblock In \emph{Proceedings of the 2019 Conference of the North {A}merican
  Chapter of the Association for Computational Linguistics: Human Language
  Technologies, Volume 1 (Long and Short Papers)}, pages 4171--4186,
  Minneapolis, Minnesota. Association for Computational Linguistics.

\bibitem[{Feng et~al.(2019)Feng, Li, and Hoey}]{feng2019keep}
Steven~Y. Feng, Aaron~W. Li, and Jesse Hoey. 2019.
\newblock \href {https://doi.org/10.18653/v1/D19-1272} {Keep calm and switch
  on! preserving sentiment and fluency in semantic text exchange}.
\newblock In \emph{Proceedings of the 2019 Conference on Empirical Methods in
  Natural Language Processing and the 9th International Joint Conference on
  Natural Language Processing (EMNLP-IJCNLP)}, pages 2701--2711, Hong Kong,
  China. Association for Computational Linguistics.

\bibitem[{Garbe(2019)}]{symspell2019}
Wolf Garbe. 2019.
\newblock Symspell.
\newblock \url{https://github.com/wolfgarbe/SymSpell}.

\bibitem[{Glockner et~al.(2018)Glockner, Shwartz, and
  Goldberg}]{glockner2018breaking}
Max Glockner, Vered Shwartz, and Yoav Goldberg. 2018.
\newblock {B}reaking {NLI} systems with {S}entences that {R}equire {S}imple
  {L}exical {I}nferences.
\newblock In \emph{Proceedings of the 56th Annual Meeting of the Association
  for Computational Linguistics (Volume 2: Short Papers)}, volume~2, pages
  650--655.

\bibitem[{Holtzman et~al.(2019)Holtzman, Buys, Du, Forbes, and
  Choi}]{holtzman2019curious}
Ari Holtzman, Jan Buys, Li~Du, Maxwell Forbes, and Yejin Choi. 2019.
\newblock The curious case of neural text degeneration.
\newblock \emph{arXiv preprint arXiv:1904.09751}.

\bibitem[{Jia and Liang(2017)}]{jia2017adversarial}
Robin Jia and Percy Liang. 2017.
\newblock \href {https://doi.org/10.18653/v1/D17-1215} {Adversarial examples
  for evaluating reading comprehension systems}.
\newblock In \emph{Proceedings of the 2017 Conference on Empirical Methods in
  Natural Language Processing}, pages 2021--2031, Copenhagen, Denmark.
  Association for Computational Linguistics.

\bibitem[{Kang et~al.(2019)Kang, Gangal, and Hovy}]{kang-etal-2019-male}
Dongyeop Kang, Varun Gangal, and Eduard Hovy. 2019.
\newblock \href {https://doi.org/10.18653/v1/D19-1179} {(male, bachelor) and
  (female, {P}h.{D}) have different connotations: Parallelly annotated
  stylistic language dataset with multiple personas}.
\newblock In \emph{Proceedings of the 2019 Conference on Empirical Methods in
  Natural Language Processing and the 9th International Joint Conference on
  Natural Language Processing (EMNLP-IJCNLP)}, pages 1696--1706, Hong Kong,
  China. Association for Computational Linguistics.

\bibitem[{Kang et~al.(2018)Kang, Khot, Sabharwal, and Hovy}]{kang2018adventure}
Dongyeop Kang, Tushar Khot, Ashish Sabharwal, and Eduard Hovy. 2018.
\newblock Adventure: {A}dversarial {T}raining for {T}extual {E}ntailment with
  knowledge-guided examples.
\newblock In \emph{Proceedings of the 56th Annual Meeting of the Association
  for Computational Linguistics (Volume 1: Long Papers)}, volume~1, pages
  2418--2428.

\bibitem[{Kann et~al.(2018)Kann, Rothe, and
  Filippova}]{kann-etal-2018-sentence}
Katharina Kann, Sascha Rothe, and Katja Filippova. 2018.
\newblock \href {https://doi.org/10.18653/v1/K18-1031} {Sentence-level fluency
  evaluation: References help, but can be spared!}
\newblock In \emph{Proceedings of the 22nd Conference on Computational Natural
  Language Learning}, pages 313--323, Brussels, Belgium. Association for
  Computational Linguistics.

\bibitem[{Kingma and Ba(2015)}]{kingma2014adam}
Diederik~P. Kingma and Jimmy Ba. 2015.
\newblock \href {http://arxiv.org/abs/1412.6980} {Adam: {A} method for
  stochastic optimization}.
\newblock In \emph{3rd International Conference on Learning Representations,
  {ICLR} 2015, San Diego, CA, USA, May 7-9, 2015, Conference Track
  Proceedings}.

\bibitem[{Kobayashi(2018)}]{kobayashi-2018-contextual}
Sosuke Kobayashi. 2018.
\newblock \href {https://doi.org/10.18653/v1/N18-2072} {Contextual
  augmentation: Data augmentation by words with paradigmatic relations}.
\newblock In \emph{Proceedings of the 2018 Conference of the North {A}merican
  Chapter of the Association for Computational Linguistics: Human Language
  Technologies, Volume 2 (Short Papers)}, pages 452--457, New Orleans,
  Louisiana. Association for Computational Linguistics.

\bibitem[{Kumar et~al.(2020)Kumar, Choudhary, and Cho}]{kumar2020data}
Varun Kumar, Ashutosh Choudhary, and Eunah Cho. 2020.
\newblock Data augmentation using pre-trained transformer models.
\newblock \emph{arXiv preprint arXiv:2003.02245}.

\bibitem[{Li et~al.(2020)Li, Jiang, Feng, Sprague, Zhou, and
  Hoey}]{li2020aloha}
Aaron~W Li, Veronica Jiang, Steven~Y. Feng, Julia Sprague, Wei Zhou, and Jesse
  Hoey. 2020.
\newblock \href {https://doi.org/10.1609/aaai.v34i05.6328} {Aloha: Artificial
  learning of human attributes for dialogue agents.}
\newblock In \emph{Proceedings of Thirty-Fourth AAAI Conference on Artificial
  Intelligence (AAAI-20)}, pages 8155--8163.

\bibitem[{Li et~al.(2016)Li, Galley, Brockett, Gao, and
  Dolan}]{li2015diversity}
Jiwei Li, Michel Galley, Chris Brockett, Jianfeng Gao, and Bill Dolan. 2016.
\newblock \href {https://doi.org/10.18653/v1/N16-1014} {A diversity-promoting
  objective function for neural conversation models}.
\newblock In \emph{Proceedings of the 2016 Conference of the North {A}merican
  Chapter of the Association for Computational Linguistics: Human Language
  Technologies}, pages 110--119, San Diego, California. Association for
  Computational Linguistics.

\bibitem[{Lu et~al.(2006)Lu, Zheng, Velivelli, and Zhai}]{lu2006}
Xinghua Lu, Bin Zheng, Atulya Velivelli, and Chengxiang Zhai. 2006.
\newblock \href {https://doi.org/10.1197/jamia.M2051} {Enhancing text
  categorization with semantic-enriched representation and training data
  augmentation}.
\newblock \emph{Journal of the American Medical Informatics Association :
  JAMIA}, 13:526--35.

\bibitem[{Maynez et~al.(2020)Maynez, Narayan, Bohnet, and
  McDonald}]{maynez2020faithfulness}
Joshua Maynez, Shashi Narayan, Bernd Bohnet, and Ryan McDonald. 2020.
\newblock On faithfulness and factuality in abstractive summarization.
\newblock \emph{arXiv preprint arXiv:2005.00661}.

\bibitem[{Miller(1995)}]{miller1998wordnet}
George~A Miller. 1995.
\newblock Wordnet: a lexical database for english.
\newblock \emph{Communications of the ACM}, 38(11):39--41.

\bibitem[{Onix()}]{stopwords}
Onix.
\newblock Onix text retrieval toolkit stopword list 1.
\newblock \url{http://www.lextek.com/manuals/onix/stopwords1.html}.

\bibitem[{Papanikolaou and Pierleoni(2020)}]{papanikolaou2020dare}
Yannis Papanikolaou and Andrea Pierleoni. 2020.
\newblock Dare: Data augmented relation extraction with gpt-2.
\newblock \emph{arXiv preprint arXiv:2004.13845}.

\bibitem[{Papineni et~al.(2002)Papineni, Roukos, Ward, and
  Zhu}]{papineni2002bleu}
Kishore Papineni, Salim Roukos, Todd Ward, and Wei-Jing Zhu. 2002.
\newblock Bleu: a method for automatic evaluation of machine translation.
\newblock In \emph{Proceedings of the 40th annual meeting on association for
  computational linguistics}, pages 311--318. Association for Computational
  Linguistics.

\bibitem[{Radford et~al.(2019)Radford, Wu, Child, Luan, Amodei, and
  Sutskever}]{radford2019language}
Alec Radford, Jeffrey Wu, Rewon Child, David Luan, Dario Amodei, and Ilya
  Sutskever. 2019.
\newblock Language models are unsupervised multitask learners.
\newblock \emph{OpenAI Blog}, 1(8):9.

\bibitem[{Rose et~al.(2010)Rose, Engel, Cramer, and Cowley}]{rose2010automatic}
Stuart Rose, Dave Engel, Nick Cramer, and Wendy Cowley. 2010.
\newblock Automatic keyword extraction from individual documents.
\newblock \emph{Text mining: applications and theory}, 1:1--20.

\bibitem[{See et~al.(2019)See, Pappu, Saxena, Yerukola, and
  Manning}]{see2019massively}
Abigail See, Aneesh Pappu, Rohun Saxena, Akhila Yerukola, and Christopher~D.
  Manning. 2019.
\newblock \href {https://doi.org/10.18653/v1/K19-1079} {Do massively pretrained
  language models make better storytellers?}
\newblock In \emph{Proceedings of the 23rd Conference on Computational Natural
  Language Learning (CoNLL)}, pages 843--861, Hong Kong, China. Association for
  Computational Linguistics.

\bibitem[{Socher et~al.(2013)Socher, Perelygin, Wu, Chuang, Manning, Ng, and
  Potts}]{socher2013recursive}
Richard Socher, Alex Perelygin, Jean Wu, Jason Chuang, Christopher~D Manning,
  Andrew~Y Ng, and Christopher Potts. 2013.
\newblock Recursive deep models for semantic compositionality over a sentiment
  treebank.
\newblock In \emph{Proceedings of the 2013 conference on empirical methods in
  natural language processing}, pages 1631--1642.

\bibitem[{Tevet and Berant(2020)}]{tevet2020evaluating}
Guy Tevet and Jonathan Berant. 2020.
\newblock Evaluating the evaluation of diversity in natural language
  generation.
\newblock \emph{arXiv preprint arXiv:2004.02990}.

\bibitem[{Toutanova et~al.(2003)Toutanova, Klein, Manning, and
  Singer}]{stanfordPOStagger}
Kristina Toutanova, Dan Klein, Christopher~D. Manning, and Yoram Singer. 2003.
\newblock \href {https://doi.org/10.3115/1073445.1073478} {Feature-rich
  part-of-speech tagging with a cyclic dependency network}.
\newblock In \emph{Proceedings of the 2003 Conference of the North American
  Chapter of the Association for Computational Linguistics on Human Language
  Technology - Volume 1}, NAACL '03, page 173–180, USA. Association for
  Computational Linguistics.

\bibitem[{Vaswani et~al.(2017)Vaswani, Shazeer, Parmar, Uszkoreit, Jones,
  Gomez, Kaiser, and Polosukhin}]{vaswani2017attention}
Ashish Vaswani, Noam Shazeer, Niki Parmar, Jakob Uszkoreit, Llion Jones,
  Aidan~N Gomez, {\L}ukasz Kaiser, and Illia Polosukhin. 2017.
\newblock Attention is all you need.
\newblock In \emph{Advances in Neural Information Processing Systems}, pages
  5998--6008.

\bibitem[{Wei and Zou(2019)}]{wei2019eda}
Jason Wei and Kai Zou. 2019.
\newblock \href {https://doi.org/10.18653/v1/D19-1670} {{EDA}: Easy data
  augmentation techniques for boosting performance on text classification
  tasks}.
\newblock In \emph{Proceedings of the 2019 Conference on Empirical Methods in
  Natural Language Processing and the 9th International Joint Conference on
  Natural Language Processing (EMNLP-IJCNLP)}, pages 6382--6388, Hong Kong,
  China. Association for Computational Linguistics.

\bibitem[{Wieting and Gimpel(2017)}]{wieting-gimpel-2017-revisiting}
John Wieting and Kevin Gimpel. 2017.
\newblock \href {https://doi.org/10.18653/v1/P17-1190} {Revisiting recurrent
  networks for paraphrastic sentence embeddings}.
\newblock In \emph{Proceedings of the 55th Annual Meeting of the Association
  for Computational Linguistics (Volume 1: Long Papers)}, pages 2078--2088,
  Vancouver, Canada. Association for Computational Linguistics.

\bibitem[{Wolf et~al.(2019)Wolf, Debut, Sanh, Chaumond, Delangue, Moi, Cistac,
  Rault, Louf, Funtowicz, and Brew}]{Wolf2019HuggingFacesTS}
Thomas Wolf, Lysandre Debut, Victor Sanh, Julien Chaumond, Clement Delangue,
  Anthony Moi, Pierric Cistac, Tim Rault, R'emi Louf, Morgan Funtowicz, and
  Jamie Brew. 2019.
\newblock Huggingface's transformers: State-of-the-art natural language
  processing.
\newblock \emph{ArXiv}, abs/1910.03771.

\bibitem[{Yelp()}]{YelpReviewsDataset}
Yelp.
\newblock Yelp open dataset.
\newblock \url{https://www.yelp.com/dataset }.

\bibitem[{Zhang et~al.(2019{\natexlab{a}})Zhang, Kishore, Wu, Weinberger, and
  Artzi}]{zhang2019bertscore}
Tianyi Zhang, Varsha Kishore, Felix Wu, Kilian~Q Weinberger, and Yoav Artzi.
  2019{\natexlab{a}}.
\newblock Bertscore: Evaluating text generation with bert.
\newblock \emph{arXiv preprint arXiv:1904.09675}.

\bibitem[{Zhang et~al.(2019{\natexlab{b}})Zhang, Baldridge, and
  He}]{zhang2019paws}
Yuan Zhang, Jason Baldridge, and Luheng He. 2019{\natexlab{b}}.
\newblock Paws: Paraphrase adversaries from word scrambling.
\newblock In \emph{Proceedings of the 2019 Conference of the North American
  Chapter of the Association for Computational Linguistics: Human Language
  Technologies, Volume 1 (Long and Short Papers)}, pages 1298--1308.

\bibitem[{Zhu et~al.(2018)Zhu, Lu, Zheng, Guo, Zhang, Wang, and Yu}]{selfbleu}
Yaoming Zhu, Sidi Lu, Lei Zheng, Jiaxian Guo, Weinan Zhang, Jun Wang, and Yong
  Yu. 2018.
\newblock \href {https://doi.org/10.1145/3209978.3210080} {Texygen: A
  benchmarking platform for text generation models}.
\newblock In \emph{The 41st International ACM SIGIR Conference on Research;
  Development in Information Retrieval}, SIGIR '18, page 1097–1100, New York,
  NY, USA. Association for Computing Machinery.

\end{thebibliography}
